%% file: main.tex
\pdfoutput=1
\documentclass[acmtog]{acmart}
\AtBeginDocument{%
  }

\input{macros}

\begin{document}

\setcopyright{cc}
\setcctype{by}
\acmJournal{TOG}
\acmYear{2026} \acmVolume{45} \acmNumber{4} \acmArticle{71}
\acmMonth{7} \acmDOI{10.1145/3811344}
% \maketitle

\title{\sysname: 
\rev{Native Mesh Generation with Diffusion}}

\author{Hanxiao Wang}
\authornote{Both authors contributed equally to this research.}
% \orcid{0009-0006-8926-6092}
%\orcid{1234-5678-9012-3456}
\affiliation{%
 \institution{MAIS, Institute of Automation, Chinese Academy of Sciences; School of Artificial Intelligence, University of Chinese Academy of Sciences}
  \city{Beijing}
  \country{China}}
\email{wanghanxiao18@mails.ucas.ac.cn}

\author{Ying-Tian Liu}
\authornotemark[1]
% \orcid{0000-0001-8293-3223}
\affiliation{%
  \institution{VAST}
  \city{Beijing}
  \country{China}}
\email{lytlogic@gmail.com}

\author{Yuan-Chen Guo}
% \orcid{0000-0001-6164-8343}
\affiliation{%
  \institution{VAST}
  \city{Beijing}
  \country{China}}
\email{imbennyguo@gmail.com}

\author{Qi-Yuan Feng}
% \orcid{0009-0005-7545-3308}
\affiliation{%
  \institution{Tsinghua University}
  \city{Beijing}
  \country{China}}
\email{fqy22@mails.tsinghua.edu.cn}

\author{Zi-Xin Zou}
% \orcid{0000-0003-2945-552X}
\affiliation{%
  \institution{VAST}
  \city{Beijing}
  \country{China}}
\email{zouzx1997@gmail.com}

\author{Ding Liang}
% \orcid{0000-0001-9774-4687}
\affiliation{%
  \institution{VAST}
  \city{Beijing}
  \country{China}}
\email{liangding1990@163.com}

\author{Biao Zhang}
\authornote{Corresponding authors.}
% \orcid{0000-0001-5685-6092}
\affiliation{%
  \institution{Xi'an Jiaotong University}
  \city{Xi'an}
  \country{China}}
\email{biao.z@outlook.com}

\author{Yan-Pei Cao}
\authornotemark[2]
% \orcid{0000-0002-0416-4374}
\affiliation{%
  \institution{VAST}
  \city{Beijing}
  \country{China}}
\email{caoyanpei@gmail.com}

\begin{abstract}

\rev{Generating high-quality triangle meshes is essential for film, gaming, and interactive 3D applications. Mainstream methods rely on mesh serialization and autoregressive processes, which stuggles in effective inference and is sensitive to error accumulation. In this paper, }we present \textbf{\sysname}, a \rev{diffusion method} that achieves \rev{holistic} mesh generation \rev{via decoupled} vertex and topology \rev{generation}.
%Our approach operates in two stages. 
First, we view mesh vertices as sparse voxels organized as an octree and adopt a diffusion model to generate the vertices in a coarse-to-fine manner. Second, for topology modeling, we propose \textit{Spacetime Interval}, as an extension of \textit{Spacetime Distance} to encode arbitrary edge and face topology into continuous per-vertex embeddings. It allows for a global and efficient recovery of complex topology. We then employ a diffusion model to generate the continuous embeddings on the generated vertices. Extensive experiments on the Objaverse and Toys4K datasets and in-the-wild images demonstrate that our method outperforms state-of-the-art autoregressive and two-stage baselines, effectively circumventing the inherent limitations of sequential mesh modeling. \rev{A blind user study from 3D practitioners confirms strong perceptual preference for our results.}

 % , and Famous 

\end{abstract}



\begin{teaserfigure}
  \includegraphics[width=\textwidth]{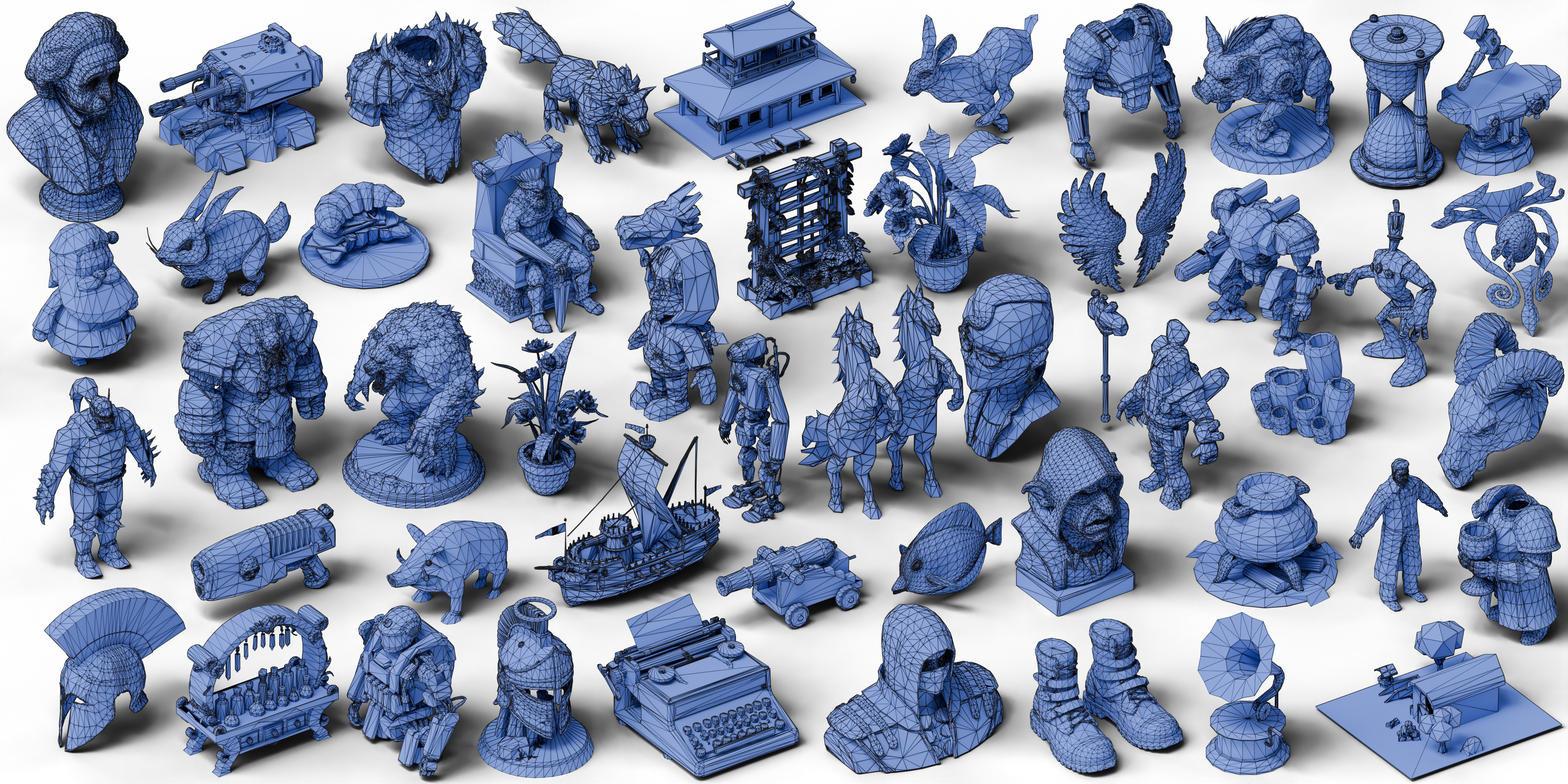}
  \vspace{-20pt}
  \caption{\textbf{Meshes generated by \sysname.} By combining hierarchical octree diffusion with our novel spacetime topology diffusion, \sysname eliminates the need for vertex sorting, achieving robust and scalable 3D meshes generation.}
  \Description{teaser}
  \label{fig:teaser}
\end{teaserfigure}


\maketitle
\section{Introduction}

Meshes play a fundamental role in modern film and gaming industries for their expressiveness, compactness and compatibility. Recently, the community is thrilled to explore high-quality mesh generation by treating meshes as 3D field~\cite{clay,triposg,trellis,lattice}. \rev{On the other hand}, artistic meshes with good topologies are still favored for manipulation, articulation and animation. Recent advancements in artistic mesh generation~\cite{siddiqui2024meshgpt,chen2024meshanything,chen2025meshanything,hao2024meshtron,tang2024edgerunner} usually represent 3D meshes with 1D discrete tokens and adopt autoregressive transformers to predict the tokens one by one. They mainly focus on how to compress meshes in shorter sequences to ease the training and inference by exploring geometry-based encoding process~\cite{chen2025meshanything,weng2025scaling,tang2024edgerunner,lionar2025treemeshgpt}. 

Despite the promising results they achieve, there remain challenges to make these approaches grounded. First, the mesh sequences usually grow linearly as the mesh complexity increases. It will take numerous auto-regressive steps to generate a single mesh. The generation time can be in minute-level \rev{despite being powered by} state-of-the-art auto-regressive inference engine~\cite{sglang,vllm}. Second, the quality of the generated mesh is significantly influenced by error accumulation in the auto-regressive process. Once some tokens are generated incorrectly, the subsequent mesh sequences would become meaningless. Moreover, from an intuitive perspective, a mesh does not bear any intrinsic order in nature, which means that it should be treated as a whole just as other signals, such as images, point clouds, etc. Thus, these key problems prompt us to think whether there exists an alternative approach to generate the 3D mesh in a holistic manner. \rev{Some current diffusion-based methods treats triangle faces in raw space~\cite{alliegro2023polydiffgenerating3dpolygonal} or latent space~\cite{he2025meshcraft}. But they still struggles in modeling the distribution of a token set in variable length, which hinders flexible scaling. } The fundamental challenge is obvious: is it possible to generate an artistic mesh as a holistic structure, \rev{while bypassing the difficulty in modeling a token set in variable length}?

\rev{To overcome the challenge,} we present \textbf{\sysname}, a novel framework that eliminates serialization entirely by leveraging diffusion processes for both geometry and topology. Our approach disentangles mesh generation into two parallelizable stages: hierarchical vertex generation and vertex-conditioned topology generation.

For geometry, we abandon the 1D sequential paradigm in favor of a spatial hierarchical representation. We treat vertices as sparse occupancy signals within an octree~\cite{octree} and propose a coarse-to-fine vertex octree diffusion model. By defining the diffusion process directly over the tree structure, we generate geometry layer-by-layer, establishing the global shape and progressively adding high-frequency details.

For topology, we introduce \textit{Spacetime Interval}, which generalizes the pairwise spacetime distance from SpaceMesh~\cite{shen2024spacemesh} to a higher-order indicator. This allows us to encode not just edges, but also complex even non-manifold face topology directly into per-vertex embeddings. Instead of predicting a sequence of discrete face indices~\cite{nash2020polygen}, we propose a topology autoencoder to encode vertices into a continuous geometric latent space.  A diffusion model then generates these latents on the vertices. Leveraging the Spacetime Interval can easily extract edges and faces. Crucially, unlike previous metric-based methods restricted to manifolds, our formulation handles arbitrary artistic topology, including non-manifold geometry.

By eliminating the requirement for sorting and sequential prediction, \sysname achieves robust, scalable, and high-quality mesh generation. We validate our approach on the Objaverse and Toys4K datasets and in-the-wild images, demonstrating that \sysname outperforms state-of-the-art autoregressive and two-stage baselines in generation quality.

In summary, our contribution is as follows:
\begin{itemize}
    \item We propose \sysname, a fully diffusion-based framework for artistic mesh generation that is inherently \textbf{sort-free}, eliminating the bottlenecks of autoregressive serialization.
    \item We design a hierarchical \textbf{octree diffusion} method to generate vertex structures efficiently in a coarse-to-fine and holistic manner.
    \item We present the \textbf{Spacetime Interval}, a geometric indicator designed to encode topology into per-vertex features. Utilizing this metric within a topology autoencoder, we obtain continuous latents that unlock the capability of diffusion-based topology generation.
    \item We demonstrate state-of-the-art performance on large-scale datasets, successfully generating complex meshes with arbitrary topologies while avoiding the error accumulation typical of sequence-based models.
\end{itemize}

\begin{comment}

\end{comment}

\section{Related Work}

\subsection{Learning on octree structures}
Octree representations recursively partition 3D space, serving as a highly efficient data structure in network design. 

\noindent\textbf{Discriminative tasks.} Foundational works like~\cite{wang2017cnn, wang2018adaptive} established octree-based convolutions, while recent architectures have integrated Transformers (OctFormer~\cite{Wang2023OctFormer}) and State Space Models (Point Mamba~\cite{liu2024point}) to capture global dependencies. Furthermore, octrees have proven effective in neural rendering tasks, such as Octree-GS~\cite{ren2024octree}, by enabling level-of-detail management for real-time applications. 

\noindent\textbf{Octree-based 3D Generation.} Octrees have become a critical representation due to their adaptive multi-scale capabilities. To address the fixed-resolution limitations of traditional VAEs, some works ~\cite{Xiong_2025_SGP, guo2025hyper3defficient3drepresentation, deng2025efficientautoregressiveshapegeneration, wei2025octgpt} proposed to use octrees to model the latent space of 3d objects.

We leverage the octree representation to exploit its spatial sparsity for computational efficiency and its adaptive nature, which allows the model to determine a variable number of vertices based on global geometric complexity.

\subsection{Mesh Generation}
We list recent important methods about mesh generation in Table~\ref{tab:method-comp}.

\noindent\textbf{Serialization Order.} Most mesh generation methods are autoregressive and rely heavily on a specific serialization order (i.e., traversal path) to tokenize the mesh~\cite{nash2020polygen, siddiqui2024meshgpt, chen2024meshxl, chen2024meshanything, hao2024meshtron}. Recent works~\cite{chen2025meshanything, weng2025scaling} devote significant effort to optimizing these traversal sequences to facilitate convergence. In contrast, while some diffusion-based methods~\cite{alliegro2023polydiffgenerating3dpolygonal} obviate the need for ordering, they often struggle with the discrete nature of mesh topology and require a pre-defined or fixed number of elements, limiting their ability to generate meshes with variable complexity.

\noindent\textbf{Decoupled Mesh Generation.} Beyond serialization, the strategy for modeling vertex coordinates and face topology is another critical design choice. Most contemporary methods~\cite{siddiqui2024meshgpt, chen2024meshxl, chen2024meshanything} adopt a unified, face-by-face generation paradigm, where vertices and their connectivity are predicted simultaneously. However, this coupled approach often leads to geometric inconsistencies, such as "cracks". Early seminal work like PolyGen~\cite{nash2020polygen} proposed a decoupled scheme that first generates the entire vertex set and then predicts the topology conditioned on these vertices. We follow this two-stage decomposition as it provides a more global perspective on the mesh structure. Distinct from PolyGen’s autoregressive formulation, we implement a full diffusion-based framework for both vertex and topology generation, leveraging the generative power of diffusion models to produce high-quality meshes with complex and coherent structures.

  \begin{table}[tb]
      \centering
      \refstepcounter{table}
      \label{tab:method-comp}
      \includegraphics{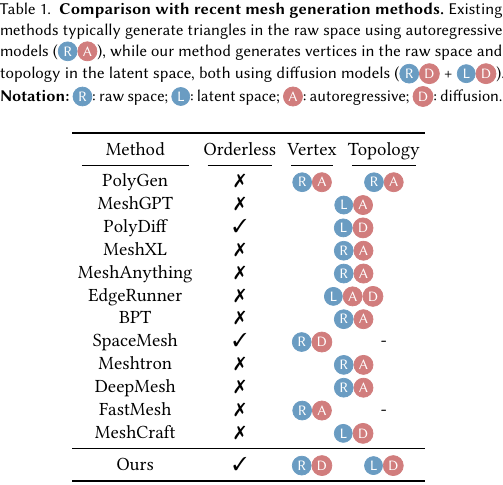}
  \end{table}

\section{Methodology}

We present \textbf{\sysname}, a generative framework designed to synthesize triangle meshes $\mathcal{M}$ conditioned on an input $\mathcal{C}$ (e.g., point clouds or images). 

Formally, we denote a mesh as $\mathcal{M}=(\mathcal{V}, \mathcal{F})$, where $\mathcal{V}$ denotes the set of vertices and $\mathcal{F}$ denotes the faces. Our approach models the probability model $p(\mathcal{M}|\mathcal{C})$. The process is decomposed into two stages: vertex generation $p(\mathcal{V}|\mathcal{C})$, and vertex-conditioned topology generation $p(\mathcal{F}|\mathcal{V}, \mathcal{C})$, formulated as:
\begin{equation}
    p(\mathcal{M}|\mathcal{C}) = 
    \underbrace{p(\mathcal{V}|\mathcal{C})}
    _{\text{vertices}} \cdot
    \underbrace{
        p(\mathcal{F}|\mathcal{V}, \mathcal{C})
    }_{\text{topology}}.
\end{equation}

\begin{figure*}[htbp]
  \includegraphics[width=\linewidth]{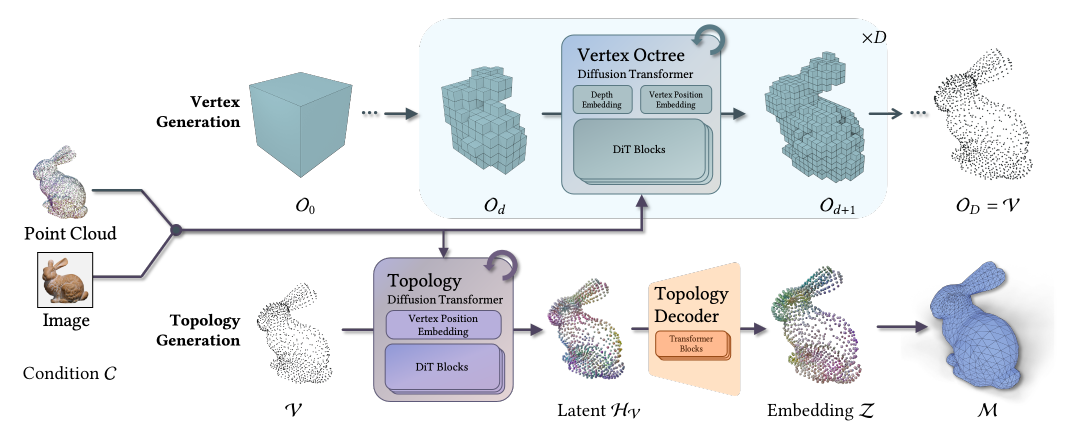}
  \vspace{-12pt}
  \caption{\textbf{Pipeline of \sysname.} Given a condition (e.g., image or point cloud), we first generate vertices via a coarse-to-fine octree diffusion model. Subsequently, we synthesize per-vertex topology latents, which are decoded into embeddings to recover mesh edges and faces via spacetime intervals.}
  \Description{pipeline}
  \label{backbone}
  % \vskip -0.3cm
    \vspace{-5pt}
\end{figure*}

\begin{table}[t]
    \caption{\textbf{Table of notations.}}
    \vspace{-6pt}
    \centering
    \renewcommand{\arraystretch}{1.1}
    \setlength{\tabcolsep}{6pt}
    \begin{tabular}{c l l}
        \toprule
        & \textbf{Symbol} & \textbf{Meaning} \\
        \midrule

        \multirow{4}{*}{\rotatebox[origin=c]{90}{Train data}}
        & $\mathcal{M}=(\mathcal{V}, \mathcal{F})$ & Mesh \\
        & $\mathcal{V}=\{\mathbf{v}\in\mathbb{R}^3\}$ & Vertices \\
        & $\mathcal{F}=\{\mathbf{f}\in \mathcal{V}\rev{^3}\}$ & Faces \\
        & $\mathcal{C}$ & Condition \\

        \midrule

        \multirow{3}{*}{\rotatebox[origin=c]{90}{Vertex}}
        & $\mathcal{O}=\{\mathbf{O}_d\}_{d=0}^{D}$ & Octrees of all depths \\
        & $\mathbf{O}_d=\{n_{d,k}\}_k$ & Octree at depth $d$ \\
        & $n_{d,k}\in\{0,1\}$ & Node $k$ at depth $d$ \\

        \midrule

        \multirow{6}{*}{\rotatebox[origin=c]{90}{Topology}}
        & $\mathcal{Z}=\{\mathbf{z}_\mathbf{v}\mid \mathbf{v}\in\mathcal{V}\}$ 
        & \multirow{2}{*}{\rev{Spacetime embeddings}} \\
        & $\mathbf{z}_\mathbf{v}=[\mathbf{s}_\mathbf{v}, \mathbf{t}_\mathbf{v}] \in\mathbb{R}^{c}$ & \\
        & $\mathbf{t}_\mathbf{v}\in\mathbb{R}^{c/2}$ & Temporal component \\
        & $\mathbf{s}_\mathbf{v}\in\mathbb{R}^{c/2}$ & Spatial component \\
        & $d_{\mathbf{u},\mathbf{v}}\in\mathbb{R}$ & Edge indicator \\
        & $d_{\mathbf{u},\mathbf{v},\mathbf{w}}\in\mathbb{R}$ & Face indicator \\

        \bottomrule
    \end{tabular}
    \vspace{-10pt}
    \label{tab:notations}
\end{table}

In Sec.~\ref{sec:method-vert-gen}, we describe how to model the vertex distribution $p(\mathcal{V}|\mathcal{C})$. The model operates in a coarse-to-fine manner: we first generate a coarse set of vertices and subsequently refine them level-by-level. Specifically, each level is modeled by a diffusion model operating on an octree representation.

In Sec.~\ref{sec:method-conn-gen}, we detail how to learn the topology distribution $p(\mathcal{F}|\mathcal{V}, \mathcal{C})$. In this stage, we train two components, (1) a topology autoencoder that encodes the topological information of edges and faces into per-vertex latent features; (2) a diffusion model trained within the latent space.

\paragraph{Comparison with Existing Methods.}
Our design addresses the fundamental limitations of autoregressive  mesh generation. 
Standard autoregressive methods~\cite{siddiqui2024meshgpt,chen2024meshxl,chen2024meshanything,chen2025meshanything,weng2025scaling} treat the mesh as an 1D sequence, which necessitates an arbitrary sorting order (e.g., Z-order). 
This formulation creates two issues: (1) sensitivity to permutation, where different sortings yield inconsistent results, and (2) error accumulation, where early mistakes lead to broken structures. 
While recent two-stage approaches like FastMesh~\cite{kim2025fastmesh} attempt to mitigate this, they still rely on sorting for autoregressive vertex generation and are limited to predicting wireframes without face information. 
In contrast, \textbf{\sysname} is a native mesh generation framework that is inherently \textit{sort-free}. 
By adapting diffusion models for both vertex synthesis and topology modeling, we eliminate the need for serialization, enabling robust generation of complete surface meshes.

\subsection{Vertex Generation}\label{sec:method-vert-gen}
We represent the vertex set $\mathcal{V}$ as an octree structure $\mathcal{O}$, denoted as
$$\mathcal{V}\rightarrow\mathcal{O}=\{\mathbf{O}_0, \mathbf{O}_1, \cdots, \mathbf{O}_D\},$$ 
where each coordinate is quantized into $D$-bit and $\mathbf{O}_d$ represents the occupancy status of nodes at depth $d$ (see Fig.~\ref{backbone}). \rev{The final vertex set $\mathcal{V}$ is obtained by extracting the 3D coordinates of all occupied leaf nodes in $\mathbf{O}_D$, i.e., the centers of the occupied grid cells at the finest octree level.} We formulate the conditional generation of vertices as a sequential probabilistic process, factorized across the tree depths:
\begin{equation}
    p(\mathcal{V} | \mathcal{C}) = p(\mathcal{O} | \mathcal{C}) =\prod_{d=1}^{D} p(\mathbf{O}_d | \mathbf{O}_{d-1}, \mathcal{C}).
\end{equation}
In other words, each level of octree model $\mathbf{O}_d$ is conditioned on the previous level $\mathbf{O}_{d-1}$.

\subsubsection{Training}
We train a unified network capable of modeling the probabilities $p(\mathbf{O}_d | \mathbf{O}_{d-1}, \mathcal{C})$ across levels simultaneously. Specifically, an octree state $\mathbf{O}_d$ at depth $d$ consists of a set of nodes $\{\mathbf{n}_{d, k}\}_k$, where the value of a node $\mathbf{n}_{d, k}\in\{0, 1\}$ indicates whether a node is occupied ($1$) or empty ($0$). Given the state at the previous level $\mathbf{O}_{d-1}$, we focus solely on the occupied nodes and we aim to train a probabilistic model to predict the occupancy of their corresponding child nodes at level $d$ (see Fig.~\ref{backbone}).

In our octree formulation, each node subdivides into $8$ child nodes. Consequently, the training target is represented by a binary  vector of size $|\sum_k \mathbf{n}_{d-1, k}| \times 8$ where $|\sum_k \mathbf{n}_{d-1, k}|$ gives the count of occupied nodes at the previous level $d-1$. 

We train a unified network across all depths to predict the occupancy of the $8$ child nodes for each occupied parent node. To apply continuous generative modeling, we cast the binary targets into real values $\{0, 1\}$ and employ flow matching~\cite{flowmatching,rectifiedflow} with velocity prediction parameterization. 
The backbone is a diffusion transformer that treats the $8$-value occupancy pattern as a single token. We utilize 3D RoPE~\cite{rope} for position encoding and a learnable depth embedding before all transformer blocks. 
The input condition $\mathcal{C}$, is encoded using jointly trained VecSet~\cite{zhang20233dshape2vecset3dshaperepresentation} encoder (for point clouds) or pretrained DINOv3~\cite{dinov3} (for images), and integrated into the denoising network with cross-attention mechanisms (see Fig.~\ref{backbone}).
\subsubsection{Inference}
The inference process operates in a coarse-to-fine manner. We start with the root level octree $\mathbf{O}_0$ containing a single occupied node.
For subsequent depths, we iteratively extract the occupied nodes from the formerly predicted or given (for the root) $\mathbf{O}_{d-1}$ and employ the trained diffusion model to predict the occupancy of the next level $\mathbf{O}_{d}$ conditioned on the same input $\mathcal{C}$.

\subsection{Topology Generation}\label{sec:method-conn-gen}

\begin{figure}[tbp]
  \centering
  \includegraphics[width=\linewidth]{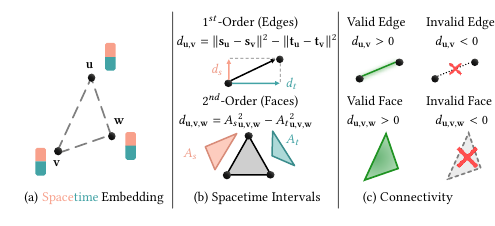}
    \vspace{-15pt}
  \caption{\textbf{Illustration of topological indicators.} We define connectivity using spacetime intervals. The 1st-order interval (top) determines edge existence based on squared distance differences, while the 2nd-order interval (bottom) determines face validity based on the difference between temporal and spatial squared areas.}
  \Description{metric}
  \vspace{-10pt}
  \label{metric}
  % \vskip -0.3cm
\end{figure}

To enable the generation of mesh topology, we design a topology autoencoder and employ a diffusion model conditioned on the vertices, formulated as $$p(\mathcal{F}|\mathcal{V}, \mathcal{C}).$$
\rev{We model edge and face topology separately. This decomposition is motivated by computational efficiency: directly recovering faces from all vertex triplets requires $O(|\mathcal{V}|^3)$ computation, whereas our edge-first strategy reduces this to $O(|\mathcal{V}|^2)$ for edge recovery followed by triangle enumeration on the sparse edge graph.}
We encode the mesh topology (i.e. edges and faces) into latent representations. Specifically, we learn per-vertex embeddings to infer whether two vertices form an edge and three vertices form a face (illustrated in Fig.~\ref{metric}). 
Let $\mathcal{Z}=\{\mathbf{z_v}\}_{\mathbf{v}\in\mathcal{V}}$ denote the per-vertex embeddings. We begin with modeling edge connectivity and extend it to faces later.

\paragraph{Edge Topology.} Naive approach is to measure some distance between two vertex embeddings,  like Euclidean distance or cosine similarity. Taking Euclidean distance as an example, 
$$d^{\text{Ecld}}_{\mathbf{u}, \mathbf{v}}=\|\mathbf{z_u}-\mathbf{z_v}\|^2.$$
the existence of an edge connecting $\mathbf{u}, \mathbf{v}$ can be determined by whether the distance $d^{\text{Ecld}}_{\mathbf{u}, \mathbf{v}}$ is greater than a threshold. To enable the distance to represent complex connectivity,  SpaceMesh~\cite{shen2024spacemesh} proposes a spacetime distance that converges much faster than the two naive choices, \begin{equation}\label{eq:hyper-quadrance}
    d_{\mathbf{u}, \mathbf{v}}=\|\mathbf{s_u}-\mathbf{s_v}\|^2-\|\mathbf{t_u}-\mathbf{t_v}\|^2,\quad \text{where }\mathbf{z_u}=[\mathbf{s_u}, \mathbf{t_u}], \mathbf{z_v}=[\mathbf{s_v}, \mathbf{t_v}].
\end{equation}
Here $[ ]$ denotes concatenation. \rev{$\mathbf{s_v} \in \mathbb{R}^{c/2}$ and $\mathbf{t_v} \in \mathbb{R}^{c/2}$ denote the spatial and temporal subvectors of the vertex embedding $\mathbf{z_v}$, respectively.} $d_{\mathbf{u}, \mathbf{v}}>0$ corresponds to a space-like separation, which we interpret as the existence of an edge between the two vertices. Conversely, a time-like separation implies no connection. We refer $d_{\mathbf{u}, \mathbf{v}}$ as \textbf{1st Order Spacetime Interval}.

\begin{revblock}
There is a strong theoretical motivation for favoring spacetime distance over standard Euclidean distance. Euclidean metrics inherently impose a form of transitive proximity: if a point is close to two other points, those two points are also encouraged to be close to each other. This implicit clustering bias makes it difficult to faithfully represent complex topologies, where nodes may share local neighborhoods without being mutually connected. Our spacetime formulation bypasses this bottleneck by utilizing an indefinite metric signature (Minkowski space). The temporal component acts as a learned slack variable to dynamically absorb such topological distortions. Intuitively, the subtractive nature of the spacetime interval ($\|\mathbf{s}\|^2 - \|\mathbf{t}\|^2$) provides higher expressive power than purely positive-definite metrics, enabling the network to cleanly isolate unconnected vertex pairs that merely appear close in the ambient space. This perspective is consistent with the viewpoint presented in SpaceMesh~\cite{shen2024spacemesh}. Furthermore, this theoretical advantage is strongly validated by our ablation study (Tab.~\ref{tab:ablation_results}), where confining the latent space to Euclidean ($L_2$) or spherical metrics causes massive degradation in topological accuracy.
\end{revblock}

\paragraph{Face Topology.} Although SpaceMesh~\cite{shen2024spacemesh} proposes an alternative representation for faces, its face recovery relies heavily on half-edge data structures, restricting its application to manifold meshes. To accommodate arbitrary topologies, we propose a novel indicator for face connectivity that extends the spacetime concept to higher dimensions. Analogous to edge modeling, a straightforward Euclidean indicator for face existence is the squared area of the triangle formed by three vertices. Specifically, let $\mathbf{z_{uw}}=\mathbf{z_u}-\mathbf{z_w}$, $\mathbf{z_{vw}}=\mathbf{z_v}-\mathbf{z_w}$. We define
\begin{equation}
    d^{\text{Ecld}}_{\mathbf{u}, \mathbf{v}, \mathbf{w}} = \|\mathbf{z_{uw}}\|^2\|\mathbf{z_{vw}}\|^2- \langle\mathbf{z_{uw}},\mathbf{z_{vw}}\rangle^2 \propto A^2_{\mathbf{u},\mathbf{v},\mathbf{w}}
\end{equation}
which represents the squared area of triangles (scaled by a factor of $4$). Inspired by the 1st Order Interval in Eq.~\ref{eq:hyper-quadrance}, we propose the spacetime counterpart as:
\begin{equation}\label{eq:hyper-quadrea}
d_{\mathbf{u}, \mathbf{v}, \mathbf{w}} = {A_s}^2_{\mathbf{u},\mathbf{v},\mathbf{w}} - {A_t}^2_{\mathbf{u},\mathbf{v},\mathbf{w}}
\end{equation}
which is the interval of squared areas between temporal and spatial components. We refer $d_{\mathbf{u}, \mathbf{v}, \mathbf{w}}$ as \textbf{2nd Order Spacetime Interval}.

\subsubsection{Training} 
\begin{figure}[tbp]

  \centering
  \includegraphics[width=\linewidth]{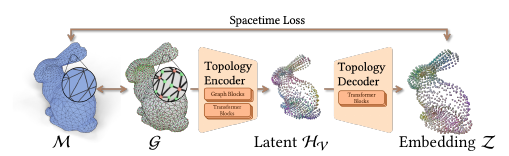}
  
  \caption{\textbf{Architecture of the Topology VAE.} The input mesh $\mathcal{M}$ is first converted into a graph $\mathcal{G}$. The encoder, consisting of interleaved Graph and Transformer blocks, captures both local and global geometric information to produce per-vertex latents $\mathcal{H}_\mathcal{V}$. The decoder then maps these latents to spacetime embeddings $\mathcal{Z}$, supervised by the Spacetime Loss.}
  \Description{Topology VAE}
  \label{topovae}
  \vspace{-10pt}
  % \vskip -0.3cm
\end{figure}

We train two networks: (1) a topology autoencoder encodes topological information into per-vertex latents; (2) a diffusion model that generates these latents conditioned on the input $\mathcal{C}$ and the vertex set $\mathcal{V}$ generated from the previous stage.
\paragraph{Topology Autoencoder} Formally, the KL-regularized autoencoder are defined as %\lyt{TODO: distinguish the vertex latent and vertex embedding},
\begin{equation}
    \begin{aligned}
    h &= \text{Enc}(\mathcal{V}, \mathcal{F}) && \text{encoder} \\
    \mu, \sigma &= \text{Linear}(h) && \smash{\lower9pt\hbox{KL bottleneck}} \\ %\smash{\lower{10pt}\text{KL bottleneck}} \\
    \mathcal{H}_\mathcal{V} &= \mu + \sigma \epsilon, \epsilon \sim \mathcal{N}(0, \mathcal{I}) && \\
    \mathcal{Z} &= \text{Dec}(\mathcal{H}_\mathcal{V}) && \text{decoder}
    \end{aligned}
\end{equation}
As shown in Fig.~\ref{topovae},
in the encoding process, we treat each mesh as a graph $\mathcal{G}$, with vertices and faces as the nodes and connections between incident vertices and faces as the edges.  To process the graph topology, we employ a hybrid architecture with interleaved graph convolutions and standard transformer blocks. \rev{Specifically, the encoder consists of interleaved SAGEConv~\cite{DBLP:journals/corr/HamiltonYL17} graph convolution layers and standard Transformer blocks. The graph layers encode topology connection effectively, while the Transformer blocks enable global information exchange across disconnected mesh components.} The input node features are the vertex positions and face centroids.

The latent bottleneck is regularized by KL divergence~\cite{vae}. The decoder consists of a stack of transformer blocks operating only on the vertex nodes and a linear head projecting the transformer output to per-vertex embeddings. In practice, we adopt separate embeddings for edge and face indicators. Then we sample vertex pairs or triplets to calculate their edge or face indicators in Eq.~\ref{eq:hyper-quadrance} and Eq.~\ref{eq:hyper-quadrea}, and apply binary cross entropy loss on it. 
Since the edges and faces are extremely sparse in vertex pairs and triplets, we employ a simple loss balancing strategy to achieve better training stability and convergence. For edges, we divide all vertex pairs into TP, TN, FP, FN (T-true, F-false, P-positive, N-negative) groups and average the binary cross entropy losses within each group before aggregating them: 
\begin{equation}
\mathcal{L}_{\text{edge}} = \frac{1}{4}\Sigma_{g\in\{\text{TP,TN,FP,FN\}}}\mathbb{E}_{(\mathbf{v_i}, \mathbf{v_j})\in g}[\text{BCE}(d_{\mathbf{v_i}, \mathbf{v_j}}, e_{ij})]
\label{loss}
\end{equation}
 where $e_{ij}=1$ indicates that there exists an edge $(\mathbf{v_i}, \mathbf{v_j})$ and otherwise $e_{ij} = 0$. At each step, all vertex pairs are supervised with edge labels. For faces, we adopt a similar balanced loss except that we supervise all positive triplets and sampled negative triplets because vertex triplets are numerous. 

\paragraph{Topology Generative Model} We train a diffusion model to generate the per-vertex latents $\mathcal{H}_\mathcal{V}$ conditioned on the vertex set $\mathcal{V}$ and the input condition $\mathcal{C}$, employing flow matching~\cite{flowmatching,rectifiedflow} with velocity prediction parameterization. The vertex position is incorporated via 3D RoPE. The input condition is injected into the main network with the same approach as the vertex generation stage.

\subsubsection{Inference}
Upon obtaining the vertex embeddings using the diffusion sampling and decoding process, we reconstruct the mesh topology through a hierarchical inference process. Since face recovering over all possible vertex triplets requires $O(|\mathcal{V}|^3)$ computation which is hard to handle, we start from edges. First, we recover the edge set by computing the pairwise spacetime interval for all vertex pairs. Then we identify \emph{candidate faces} by searching for all closed 3-cycles (i.e., triangles) formed by the recovered edges and verify each candidate $(\mathbf{u}, \mathbf{v}, \mathbf{w})$ using the 2nd order spacetime interval. This ``edges-first'' strategy ensures topologically consistency while significantly reducing computational overhead. \rev{By construction, a face can only be validated on edges that have already been confirmed, preventing face-edge inconsistency and ensuring topological coherence.}

\section{Experiments}
\subsection{Implementation Details}

\paragraph{Vertex Diffusion.}
We train a diffusion transformer~\cite{dit} (DiT) with approximately 2 billion parameters to recover vertex positions conditioned on point clouds and images.
For the condition encoder, we employ a VecSet~\cite{zhang20233dshape2vecset3dshaperepresentation} architecture comprising 8 layers with a hidden dimension of 2048. The input to the encoder consists of 8192 points sampled from the mesh surface with their surface normals. These features are projected into 1024 tokens with a dimension of 2048 and injected into the DiT.
For the image encoder, we utilize a pre-trained DINOv3~\cite{dinov3} encoder to get image feature and injected into the DiT.  \rev{For point cloud conditioning, we uniformly sample 8{,}192 points from the mesh surface along with their surface normals. For image conditioning, we render images at $512 \times 512$ resolution with the camera uniformly sampled in front of the object.}
We set the depth $D=9$ which means the vertex positions are normalized and quantized into integers within the range $[0, 511]$. 
The model is trained using the Adam optimizer on 32 NVIDIA A100 GPUs for 400K steps. We use a learning rate of $1 \times 10^{-4}$ and a weight decay of $0.01$.
% Inference
For inference, we employ the DPM-Solver~\cite{dpm-solver} sampler, performing \rev{20} steps for each octree level.

\paragraph{Topology Autoencoder.}
We train a network with around 100 million parameters to encode topological information.
The encoder processes vertices and face centroids using a hybrid architecture of interleaved graph blocks (containing graph convolutions) and transformer blocks. It consists of totally 24 layers with a hidden dimension of 512. The latent bottleneck dimension is set to 64, regularized by a KL divergence loss.
The decoder employs a pure attention-based architecture with 16 layers and a hidden dimension of 1024.
The model is trained with the Adam optimizer on 32 NVIDIA A100 GPUs for 50K steps, with a learning rate of $1 \times 10^{-4}$.

\paragraph{Topology Diffusion.}
We train another DiT with 2 billion parameters to generate topology latents on the vertex conditioned on $\mathcal{C}$.
This model shares the architecture and hyperparameters as the vertex diffusion model.
The model is trained using the Adam optimizer on 32 NVIDIA A100 GPUs for 200K steps, with a learning rate of $1 \times 10^{-4}$.
Similar to the geometry stage, we use the DPM-Solver~\cite{dpm-solver} with \rev{20} inference steps. \rev{
For inference, generating a mesh with up to 20{,}000 faces takes approximately 60 seconds on a single 24GB GPU.}

\paragraph{Dataset.} 
For training of all stages, we curate a dataset from Objaverse~\cite{deitke2022objaverseuniverseannotated3d} and ObjaverseXL~\cite{deitke2023objaversexluniverse10m3d} with meshes with fewer than 20,000 faces. This results in a subset of approximately 1 million meshes. To ensure reliable evaluation, we perform testing on two diverse datasets that are excluded from the training phase:
\begin{itemize}
    \item \textbf{Objaverse:} We collect a test set of 500 models drawn randomly from the non-training portion of Objaverse.
    \item \textbf{Toys4K:} To evaluate performance on low-poly shapes, we select 900 examples from the Toys4K dataset~\cite{toys4k} with a face count threshold of 4,000.
\end{itemize}

\subsection{Point-Cloud-Conditioned Mesh Generation}

% \subsection{Mesh Generation Results}
\begin{table}[t]
 \centering
 \caption{\textbf{Quantitative comparison of mesh reconstruction quality on multiple datasets.}}
 \vspace{-6pt}
 \label{tab:combined_results_wide}
 \resizebox{0.9\linewidth}{!}{
 \begin{tabular}{lcccc}
 \toprule
 & \multicolumn{2}{c}{\textbf{Objaverse}} & \multicolumn{2}{c}{\textbf{Toys4K}} \\
 \cmidrule(lr){2-3} \cmidrule(lr){4-5}
 Method & Hausdorff $\downarrow$ & Chamfer $\downarrow$ & Hausdorff $\downarrow$ & Chamfer $\downarrow$ \\
 \midrule
 MeshAnything & 0.327 & 0.128 & 0.264 & 0.110 \\
 MeshAnythingV2 & 0.265 & 0.089 & 0.238 & 0.090 \\
 TreeMeshGPT & 0.183 & 0.055 & 0.133 & 0.047 \\
 BPT & 0.126 & 0.043 & 0.091 & 0.037 \\
FastMesh & 0.145 & 0.040 & 0.100 & 0.033 \\
 \midrule
 \textbf{Ours} &\textbf{ 0.104} &\textbf{ 0.031} & \textbf{ 0.071} & \textbf{0.027} \\

 \bottomrule
 \end{tabular}
 }
 \vspace{-10pt}
\end{table}

In this section, we compare our point-to-mesh results with state-of-the-art methods. Our evaluation includes one-stage methods such as MeshAnything v1/v2~\cite{chen2024meshanything, chen2025meshanything}, BPT~\cite{weng2025scaling}, and TreeMeshGPT~\cite{lionar2025treemeshgpt}. Furthermore, we include \rev{FastMesh~\cite{kim2025fastmesh}} as a representative baseline for two-stage pipelines. We normalize all meshes to a $[-1, 1]$ bounding box and sample point clouds for evaluation.
Quantitative evaluation is performed using Hausdorff Distance (HD) for maximum surface deviation and Chamfer Distance (CD) for average reconstruction error. \rev{Both metrics are computed using 50{,}000 points uniformly sampled from the mesh surface.}

\subsubsection{Quantitative Analysis}
We compare \sysname with state-of-the-art baselines on the Objaverse and Toys4K datasets. The results are summarized in Tab.~\ref{tab:combined_results_wide}.
Among the baselines, BPT~\cite{weng2025scaling} emerges as the most competitive one-stage autoregressive method, achieving the best Hausdorff scores among prior works. 
FastMesh~\cite{kim2025fastmesh}, representing two-stage approaches, secures better Chamfer distances than BPT but suffers from higher Hausdorff distances. This discrepancy highlights the trade-off in existing methods:  while FastMesh excels at average surface reconstruction, its deterministic connectivity prediction module often introduces severe redundant faces and geometric outliers, thereby degrading the worst-case error measured by HD.

In contrast, \sysname outperforms all baselines across both metrics on all datasets. By eliminating the autoregressive sorting bottleneck and employing diffusion for both geometry and topology, our method achieves superior global consistency while maintaining precise local details.

\rev{To provide a more comprehensive evaluation, we report additional mesh quality metrics in Tab.~\ref{tab:additional_metrics}, including Edge Chamfer Distance (ECD), Normal Consistency (NC), average vertex count (\#V), and average face count (\#F). We compute ECD by only sampling points on edges. Here Normal Consistency is defined as the average dot product of neighboring face normals, within the range [0, 2]. \sysname achieves the lowest ECD (0.0233), indicating superior edge-level geometric accuracy. Our NC value (1.5441) is higher than baselines due to the absence of face normal prediction in our pipeline, which requires a post-processing orientation correction step. FastMesh only considers edge connectivity and thus produces numerous redundant faces.
}

\begin{table}[t]
    \centering
    \caption{\rev{\textbf{Additional mesh quality metrics on the Toy4k test set.} We report Edge Chamfer Distance (ECD), Normal Consistency (NC), average vertex count (\#V), and average face count (\#F) as complementary metrics to Tab.~\ref{tab:combined_results_wide}.}}
    \vspace{-6pt}
    \label{tab:additional_metrics}
    \resizebox{\linewidth}{!}{
    \begin{tabular}{lcccc}
    \toprule
    Method & ECD $\downarrow$ & NC $\downarrow$ & \#V & \#F \\
    \midrule
    MeshAnything~\cite{chen2024meshanything}   & 0.1413 & 0.6106 &  242.5 &   431.4 \\
    MeshAnythingV2~\cite{chen2025meshanything} & 0.1129 & 0.6238 &  667.0 &  1223.6 \\
    TreeMeshGPT~\cite{lionar2025treemeshgpt}  & 0.0670 & \textbf{0.2861} &  842.6 &  1592.2 \\
    BPT~\cite{weng2025scaling}                & 0.0471 & 0.6177 &  599.7 &  1148.5 \\
    FastMesh~\cite{kim2025fastmesh}           & 0.0357 & 1.9162 &  600.0 &  4758.7 \\
    \midrule
    \textbf{Ours} & \textbf{0.0233} & 1.5441 & 1081.8 & 2287.0 \\
    \bottomrule
    \end{tabular}
    }
    \vspace{-8pt}
\end{table}

\begin{figure*}[t]
  \centering
  \includegraphics[width=0.95\linewidth]{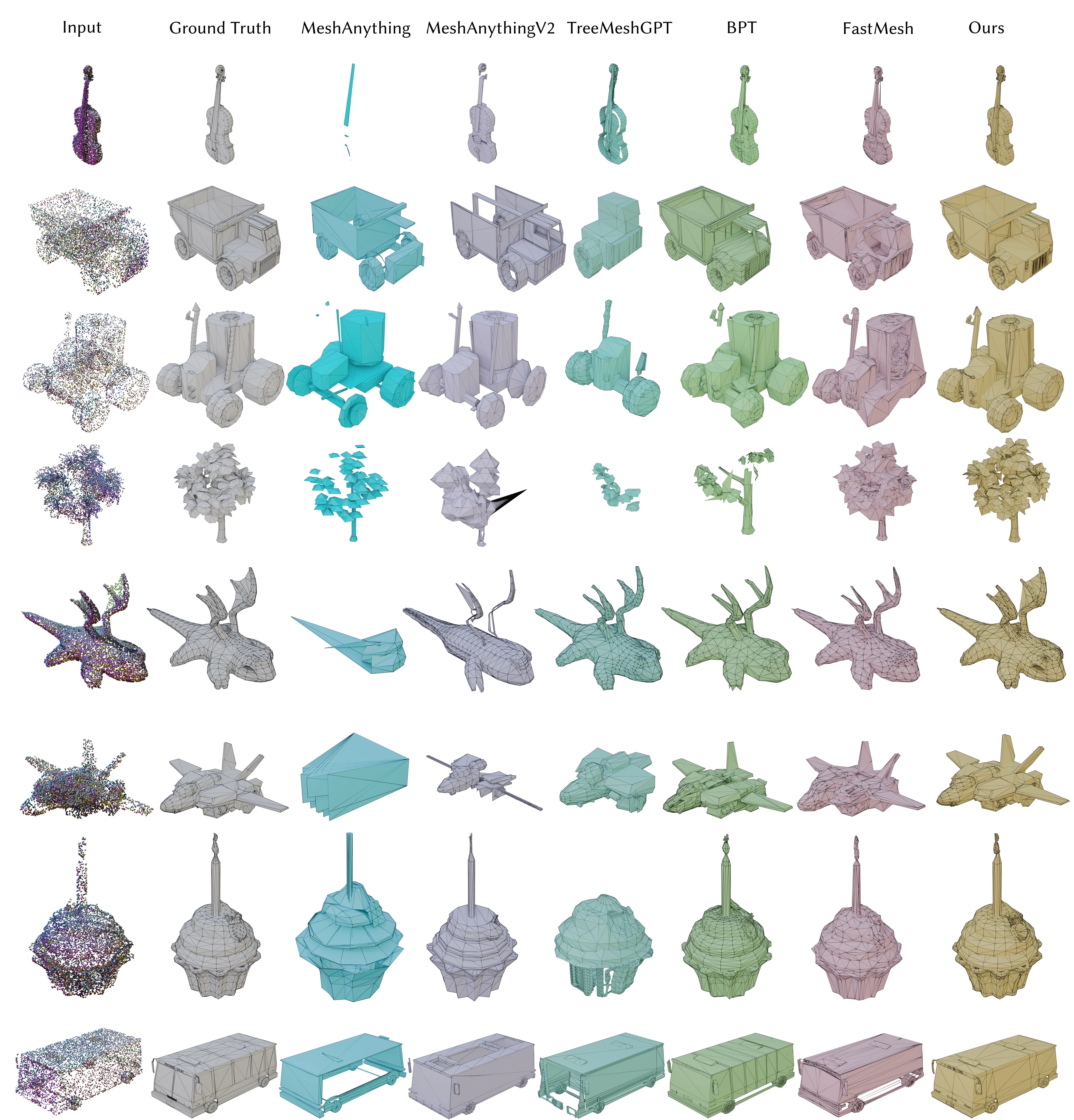}
  \vspace{-6pt}
  \caption{\textbf{Qualitative comparison of mesh reconstruction quality on the Toys4K dataset.} The leftmost column shows the input point cloud.}
  \Description{toy4k}
  \label{comparebaseline}
  \vspace{-6pt}
\end{figure*}

\subsubsection{Qualitative Comparison}
We visually compare the generated meshes in Fig.~\ref{comparebaseline}, where the results highlight specific failure modes inherent to autoregressive architectures that \sysname successfully eliminates.
FastMesh, representing two-stage models, relies on a regression network for connectivity without generative ability. This limitation often manifests as topological defects in complex regions. For instance, it generates a broken neck for the violin (row 1) and loses structural integrity in the tractor wheel (row 3), suggesting a struggle to capture the distribution of valid topologies.
Conversely, one-stage autoregressive methods like BPT are highly sensitive to sequence ordering, a weakness particularly evident in meshes with complex branching structures or enormous disconnected components. As seen in the tree model (row 4), BPT fails to generate leaves.
% likely due to the ambiguity involved in flattening a complex graph into a 1D sequence.
In contrast, \sysname demonstrates superior robustness by treating both vertices and topology in a holistic manner learned by diffusion process. Our approach generates meshes with high structural integrity and fine detail, successfully eliminating both the typical sorting artifacts from autoregressive models and the connectivity errors of deterministic regressors.

\begin{revblock}
\subsubsection{Robustness Analysis for Point-Cloud-Conditioned Generation}
We evaluate the robustness of \sysname to variations in input point cloud density and noise.

\paragraph{Input Point Cloud Density.} Tab.~\ref{tab:density_robust} evaluates robustness to varying input point cloud density. Although trained with 8{,}192 points, \sysname maintains stable performance across a 32$\times$ density range (512--16{,}384 points), with CD remaining approximately 0.023 from 2{,}048 points onward.

\begin{table}[t]
    \centering
    \caption{\rev{\textbf{Robustness to input point cloud density.} Chamfer Distance remains stable across a 32$\times$ density range (512--16{,}384 points). The model is trained with 8{,}192 points.}}
    \vspace{-6pt}
    \label{tab:density_robust}
    \resizebox{\linewidth}{!}{
    \begin{tabular}{ccccccc}
    \toprule
    \#Points & 512 & 1024 & 2048 & 4096 & 8192 & 16384 \\
    \midrule
    CD $\downarrow$ & 0.0355 & 0.0258 & 0.0235 & 0.0235 & 0.0233 & 0.0231 \\
    \bottomrule
    \end{tabular}
    }
    \vspace{-8pt}
\end{table}

\paragraph{Input Noise.} Tab.~\ref{tab:noise_robust} evaluates robustness to Gaussian noise added to the input point cloud. Under typical real-scan noise levels ($\sigma{=}0.003$), CD degrades by less than 10\% (0.0256 vs.\ 0.0233), confirming strong sensor-noise robustness.

\begin{table}[t]
    \centering
    \caption{\rev{\textbf{Robustness to input noise.} We add Gaussian noise $\mathcal{N}(0, \sigma^2)$ to the input point cloud. Under typical real-scan noise ($\sigma{=}0.003$), CD degrades by less than 10\%.}}
    \vspace{-6pt}
    \label{tab:noise_robust}
    \begin{tabular}{cccccc}
    \toprule
    $\sigma$ & 0.000 & 0.001 & 0.003 & 0.010 & 0.030 \\
    \midrule
    CD $\downarrow$ & 0.0233 & 0.0250 & 0.0256 & 0.0309 & 0.0489 \\
    \bottomrule
    \end{tabular}
    \vspace{-8pt}
\end{table}

\end{revblock}

\begin{revblock}
\subsubsection{User Study}
To evaluate the perceptual quality of generated meshes, we conduct a rigorous blind pairwise user study and collect 1{,}221 comparison results from 3D practitioners in total. Participants are required to evaluate edge flow, structural integrity, and artifact reduction. As shown in Tab.~\ref{tab:user_study}, \sysname achieves the highest Elo rating (1440.4) and pair-wise preference rate (93.0\%), significantly outperforming all baselines. This confirms that the topological quality advantage of our method is perceptually meaningful to professional users.

\begin{table}[t]
    \centering
    \caption{\rev{\textbf{User study results.} We conduct a blind pairwise study with 1{,}221 comparisons from 3D practitioners, evaluating edge flow, structural integrity, and artifact reduction.}}
    \vspace{-6pt}
    \label{tab:user_study}
    \begin{tabular}{lcc}
    \toprule
    Method & Elo Rating $\uparrow$ & Preference (\%) $\uparrow$ \\
    \midrule
    \textbf{Ours}      & \textbf{1440.4} & \textbf{93.0} \\
    FastMesh           & 1090.4 & 62.2 \\
    BPT                & 1075.3 & 60.9 \\
    TreeMeshGPT        &  928.2 & 41.2 \\
    MeshAnythingV2     &  806.6 & 28.9 \\
    MeshAnything        &  659.1 & 14.9 \\
    \bottomrule
    \end{tabular}
    \vspace{-8pt}
\end{table}
\end{revblock}

\begin{figure}[t]
  \centering
  \includegraphics[width=\linewidth]{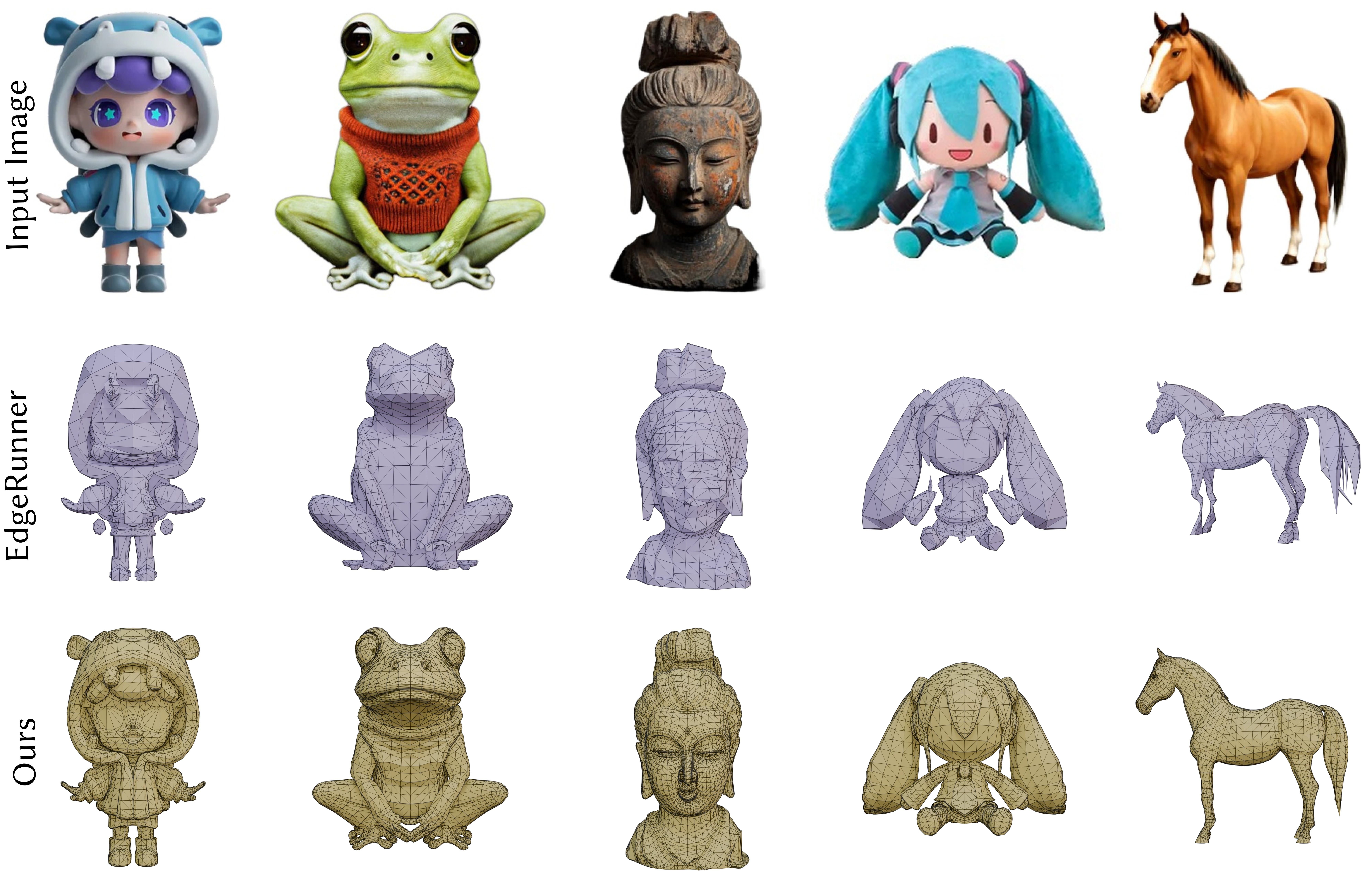}
  \vspace{-2pt}
  \caption{\textbf{Qualitative comparison with EdgeRunner on image-conditioned generation.} Results show improved geometric quality and surface detail compared to Edgerunner across diverse object categories.}
  \Description{edgerunner}
  \label{edgerunner}
  \vspace{-6pt}
\end{figure}

\subsection{Image-Conditioned Mesh Generation}

\begin{figure*}[h]
  \centering
  \includegraphics[width=0.95\linewidth]{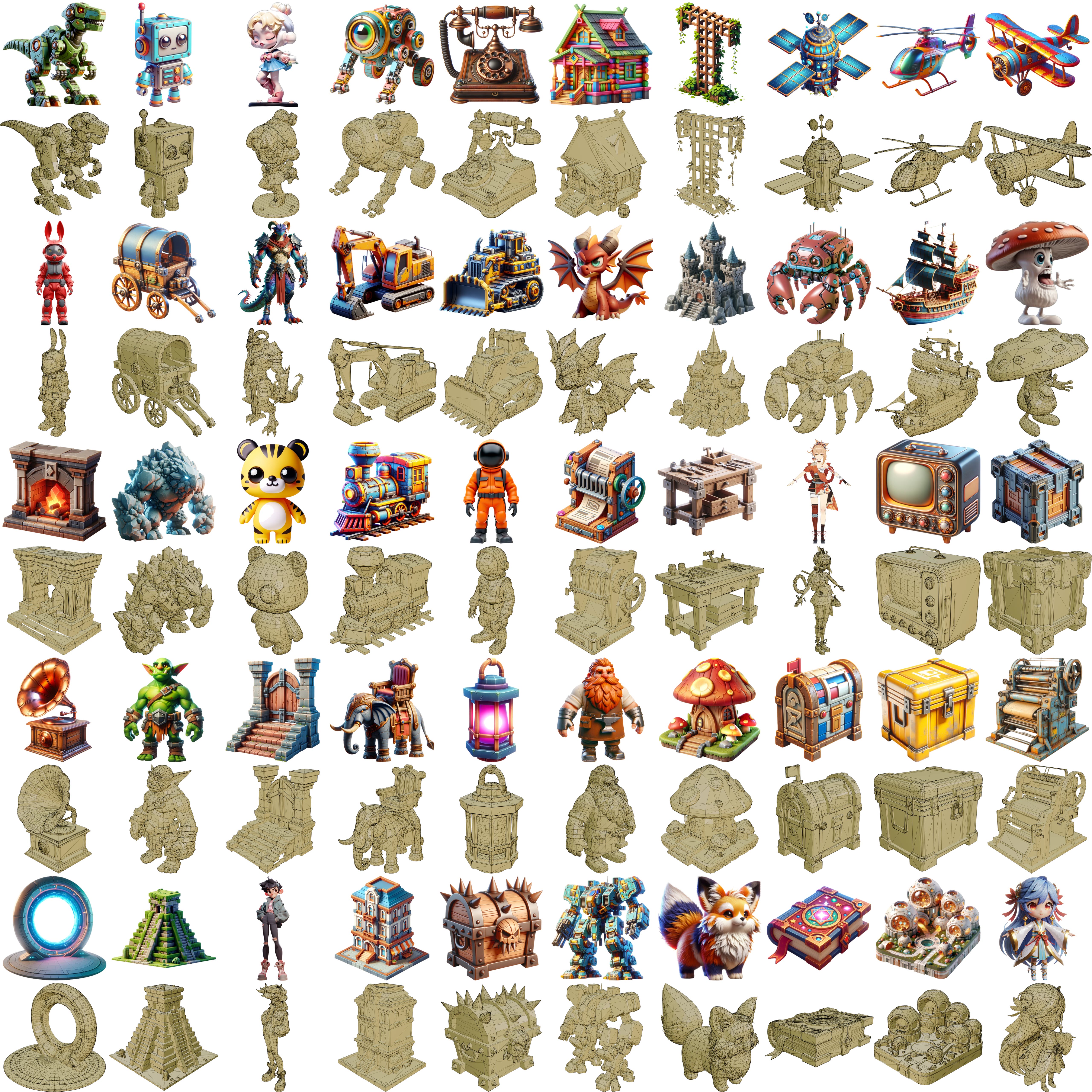}
  \vspace{-6pt}
  \caption{\rev{\textbf{Uncurated image-to-mesh generation gallery.} 50 randomly selected outputs with no manual filtering, demonstrating the consistency and robustness of \sysname across diverse categories.}}
  \Description{gallery}
  \label{fig:gallery}
  \vspace{-6pt}
\end{figure*}

We qualitatively evaluate the image conditional generation ability of our method in Fig.~\ref{edgerunner}, \rev{Fig.~\ref{trellis}, and Fig.~\ref{fig:gallery}.} 

\paragraph{Comparison with EdgeRunner}
As shown in Fig.~\ref{edgerunner}, our method significantly outperforms EdgeRunner in both geometric fidelity and topological quality. While EdgeRunner tends to produce over-smoothed, rough hulls with irregular triangulation, our approach recovers precise high-frequency details with clean element layouts. For instance, we accurately reconstruct intricate structures, such as the defined facial features of the Buddha (column 5) where the baseline yields featureless approximations. Notably, our method achieves better topological completeness. For the anime character (column 6), our method can generate a mesh with detailed hair, whereas EdgeRunner results in disconnected limbs and missing thin structures.

\begin{revblock}

\begin{figure}[t]
  \centering
  \includegraphics[width=\linewidth]{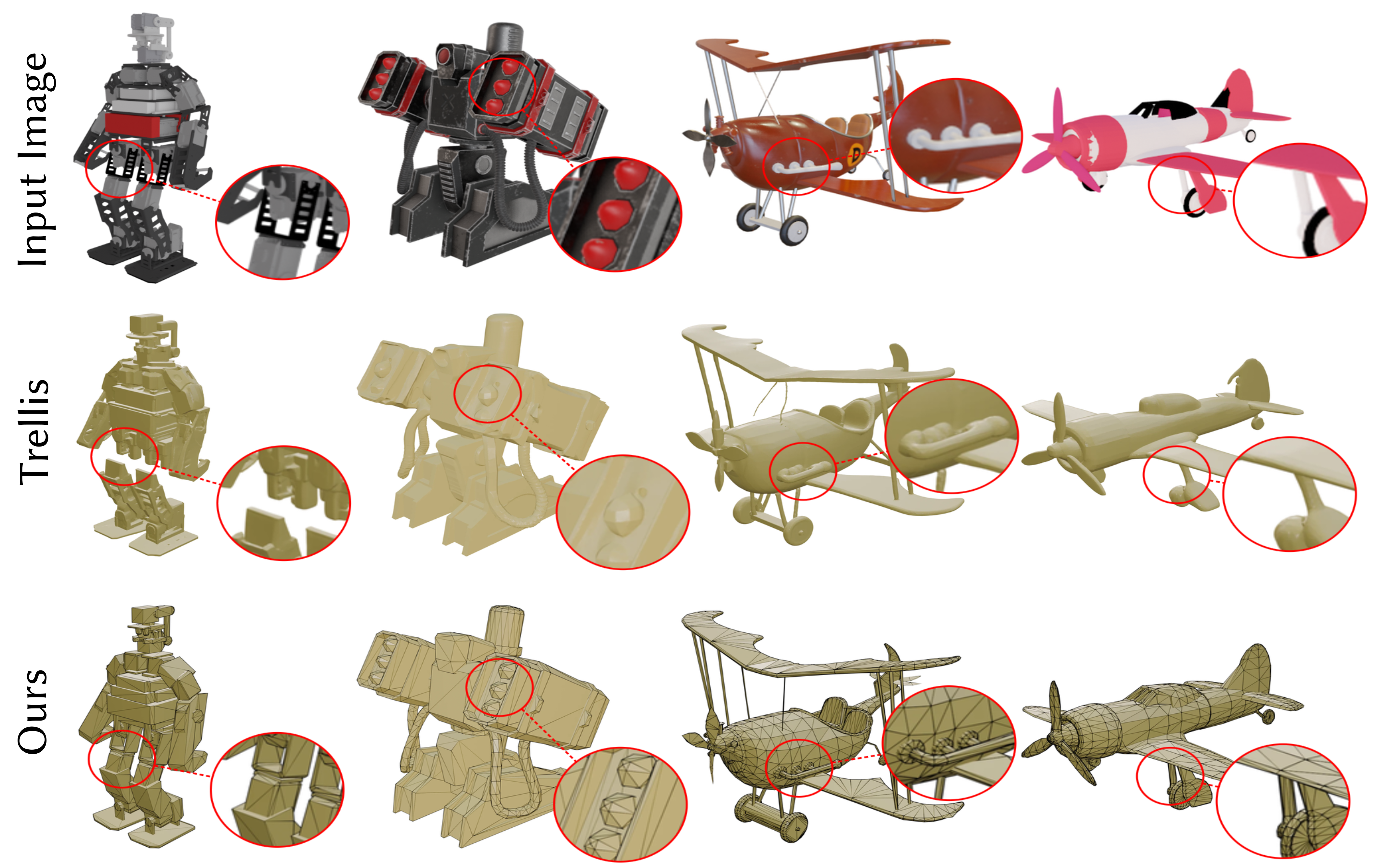}
  \vspace{-6pt}
  \caption{\rev{\textbf{Qualitative comparison with TRELLIS on image-conditioned generation.} \sysname generates native meshes with explicit topology and sharper geometric details, while TRELLIS produces smooth isosurfaces via marching cubes.}}
  \Description{trellis comparison}
  \label{trellis}
  \vspace{-6pt}
\end{figure}

\paragraph{Comparison with TRELLIS} We compare our image-conditioned generation against TRELLIS~\cite{trellis}, a state-of-the-art method based on structured latent representations and isosurface extraction. As shown in Tab.~\ref{tab:trellis_compare}, \sysname outperforms TRELLIS on both standard geometric metrics (CD: 0.0681 vs.\ 0.0802; HD: 0.2309 vs.\ 0.2503) and 3D-aware semantic similarity metrics (ULIP-Sim~\cite{ulip}, UNI3D-Sim~\cite{uni3d}). Specifically, we utilize ULIP-Sim and Uni3D-Sim to measure the cosine similarity between the generated mesh and the input image within pre-trained 3D-image joint embedding spaces. Following the evaluation protocol of recent high-fidelity 3D generation works like Lattice~\cite{lai2025latticedemocratizehighfidelity3d}, these metrics are employed to capture high-level semantic fidelity that goes beyond pure geometry.

\begin{table}[t]
    \centering
    \caption{\rev{\textbf{Image-conditioned generation: comparison with TRELLIS.} We report standard geometric metrics (CD, HD) alongside 3D-aware semantic similarity (ULIP-Sim, UNI3D-Sim).}}
    \vspace{-6pt}
    \label{tab:trellis_compare}
    \resizebox{\linewidth}{!}{
    \begin{tabular}{lcccc}
    \toprule
    Method & CD $\downarrow$ & HD $\downarrow$ & ULIP-Sim $\uparrow$ & UNI3D-Sim $\uparrow$ \\
    \midrule
    TRELLIS~\cite{trellis} & 0.0802 & 0.2503 & 0.1317 & 0.2510 \\
    \textbf{Ours}                   & \textbf{0.0681} & \textbf{0.2309} & \textbf{0.1438} & \textbf{0.2738} \\
    \bottomrule
    \end{tabular}
    }
    \vspace{-8pt}
\end{table}

Fig.~\ref{trellis} provides a qualitative comparison. While TRELLIS produces smooth surfaces via marching cubes, it tends to over-smooth fine geometric details. In contrast, \sysname preserves sharp features and generates artist-friendly mesh topology with explicit and clean edge flow.

\end{revblock}

\begin{revblock}

\paragraph{Uncurated Generation Gallery.} To demonstrate the consistency of our method beyond cherry-picked examples, Fig.~\ref{fig:gallery} shows 50 randomly selected outputs from our model conditioned on images. No manual selection or filtering was applied, emphasizing the robustness of our holistic generation paradigm across diverse categories and geometric complexities.
\end{revblock}

\subsection{Ablation Studies}
\label{sec:ablation}

\begin{table}[tbp]
    \centering
    \caption{\textbf{Quantitative comparison of face and edge metrics under different design choices}. The upper section shows results at 8k iterations. \rev{``Euclidean ($L_2$)'' and ``Spherical'' replace the spacetime interval with the corresponding metric.} The last row shows our final autoencoder at 50k iterations.}
    \vspace{-4mm}
    \label{tab:ablation_results}
    \resizebox{\linewidth}{!}{
        \begin{tabular}{lcccccc}
            \toprule
            & \multicolumn{3}{c}{\textbf{Face}} & \multicolumn{3}{c}{\textbf{Edge}} \\
            \cmidrule(lr){2-4} \cmidrule(lr){5-7}
            Experiment & Recall $\uparrow$ & Precision $\uparrow$ & F1 Score $\uparrow$ & Recall $\uparrow$ & Precision $\uparrow$ & F1 Score $\uparrow$ \\
            \midrule
            Ours & \textbf{0.99995} & \textbf{0.99976} & \textbf{0.99985} & 0.99996 & 0.99991 & 0.99994 \\
            \textit{Minkowski loss} & 0.99985 & 0.99476 & 0.99729 & \textbf{0.99998} & \textbf{0.99996} & \textbf{0.99997} \\
            \rev{\textit{Euclidean ($L_2$)}} & \rev{0.93370} & \rev{0.79619} & \rev{0.85948} & \rev{0.97632} & \rev{0.88190} & \rev{0.92671} \\
            \rev{\textit{Spherical} (cos)} & \rev{0.90626} & \rev{0.50889} & \rev{0.65178} & \rev{0.99976} & \rev{0.57707} & \rev{0.73176} \\
            \textit{w/o balance loss} & \textbf{0.99995} & 0.99956 & 0.99976 & \textbf{0.99998} & 0.99990 & 0.99994 \\
            \textit{w/o enc attn} & 0.99986 & 0.99965 & 0.99976 & \textbf{0.99998}& 0.99991 & 0.99995 \\
            \textit{w/o enc GCN} & 0.98206 & 0.94899 & 0.96522 & 0.89112 & 0.73120 & 0.80327 \\
            \textit{w/o split feature} & 0.99960 & 0.96150 & 0.98016 & 0.98796 & 0.97315 & 0.98069 \\
            \midrule
            Ours (50k) & 0.99998 & 0.99991 & 0.99994 & 1.00000 & 1.00000 & 1.00000 \\
            \bottomrule
        \end{tabular}
    }
    \vspace{-6mm}
\end{table}

To evaluate the effectiveness of some key designs, we conduct a series of ablation experiments focusing on the topology autoencoder and the proposed spacetime indicator. The quantitative results are summarized in Tab.~\ref{tab:ablation_results}.

\paragraph{Topology Indicator (\textit{Minkowski loss}).}

To evaluate the effectiveness of our proposed \textbf{2nd Order Spacetime Interval} defined in Eq.~\ref{eq:hyper-quadrea}, we investigate an alternative formulation for face topology derived from the pseudo-Euclidean inner product.
While our default \textbf{2nd Order Spacetime Interval} calculates the difference between temporal and spatial squared areas independently, the alternative indicator, which we call Minkowski Interval, couples these components within a pseudo-Euclidean inner product. Let $\mathbf{u} = \mathbf{z_u} - \mathbf{z_w}$ and $\mathbf{v} = \mathbf{z_v} - \mathbf{z_w}$ be the relative vertex embeddings. The pseudo-Euclidean inner product $[\cdot, \cdot]$ is defined as:
\begin{equation}
[\mathbf{u}, \mathbf{v}] = \langle \mathbf{s_u}, \mathbf{s_v} \rangle - \langle \mathbf{t_u}, \mathbf{t_v} \rangle,
\end{equation}
where $\langle \cdot, \cdot \rangle$ denotes the standard Euclidean inner product. The face topology indicator $d^{\text{Mink}}_{\mathbf{u}, \mathbf{v}, \mathbf{w}}$ is formulated as the determinant:
\begin{equation}
d^{\text{Mink}}_{\mathbf{u}, \mathbf{v}, \mathbf{w}} = [\mathbf{u}, \mathbf{u}][\mathbf{v}, \mathbf{v}] - [\mathbf{u}, \mathbf{v}]^2.
\end{equation}
Conceptually, this variant treats the latent space as a unified Minkowski manifold. However, as shown in Tab.~\ref{tab:ablation_results}, we observe that the baseline outperforms the Minkowski variant. Its performance on faces is significantly inferior to the original version.  We hypothesize that by decoupling the temporal and spatial geometric determinants, the model gains better numerical stability during the optimization process, as it prevents the gradients of the temporal and spatial components from interfering within the quadratic expansion of the determinant.

\begin{revblock}
\paragraph{Distance Metric (Euclidean / Spherical).} To validate the necessity of the spacetime formulation, we replace the indefinite metric with standard Euclidean ($L_2$) and Spherical (cosine similarity) metrics. For Euclidean distance, edge indicator compares $\|\mathbf{z_u}-\mathbf{z_v}\|^2$ with a learned threshold and face indicator compares $A^2_{\mathbf{u},\mathbf{v},\mathbf{w}}$ with another threshold.
For spherical distance, edge indicator uses $\cos\theta_{\mathbf{u}, \mathbf{v}} = \frac {\mathbf{z_u}\cdot \mathbf{z_v}} {\|\mathbf{z_u}\|\|\mathbf{z_v}\|}$ while face indicators uses $\frac {\cos\theta_{\mathbf{u}, \mathbf{v}}  + \cos\theta_{\mathbf{u}, \mathbf{w}}  + \cos\theta_{\mathbf{v}, \mathbf{w}} } 3$. As shown in Tab.~\ref{tab:ablation_results}, both alternatives cause massive topological degradation. The Euclidean variant achieves only 0.859 Face F1 (vs.\ 0.9999 for ours), while the Spherical variant drops further to 0.652. This confirms that the indefinite metric signature is essential: the subtractive temporal component provides the expressive slack needed to represent complex, non-planar connectivity patterns that positive-definite metrics cannot capture.
\end{revblock}

\paragraph{Loss Balancing Strategy (\textit{w/o balance loss})}
We employ a simple loss balancing strategy described in Eq.\ref{loss} due to the extreme sparsity of edges and faces in candidate vertex pairs and triplets.
Excluding this strategy (\textit{w/o balance loss}) results in a decrease in the face F1 score($0.99976$ vs. $0.99985$), validating its necessity for achieving better performance during training.

\paragraph{Graph Convolution (\textit{w/o enc GCN})}
Removing the graph convolution layers from the encoder results in the most drastic performance decline, with the edge F1 score falling to $0.88448$. This demonstrates that injecting the connectivity with graph operators is paramount. Without these layers, the model fails to effectively aggregate information from immediate neighbors, which is indispensable for understanding the initial topological structure of the mesh.

\paragraph{Full Attention (\textit{w/o enc attn})}
Restricting the encoder to only graph-based operations without global full-attention (transformer blocks) also leads to a performance decrease. This confirms that relying solely on local topological relationships is insufficient because graph convolutions cannot propagate messages between disjoint components (unconnected parts of the mesh). The global attention is therefore necessary to bridge this gap, enabling information exchange across the entire geometry regardless of connectivity.

\paragraph{Feature Decoupling Strategy (\textit{w/o feature split})}
In our default configuration, the per-vertex embedding $\mathcal{Z}$ is split into two disjoint subsets of channels: $\mathcal{Z}_{edge}$ for computing 1st-order intervals (edges) and $\mathcal{Z}_{face}$ for 2nd-order intervals (faces). In the \textit{w/o split} variant, we employ a unified embedding to represent both edge and face topology simultaneously. As shown in Tab.~\ref{tab:ablation_results}, we observe a significant performance degradation, with the face F1 score dropping from $0.99985$ to $0.98016$. This drop suggests that forcing a single feature space to satisfy the geometric constraints of both pairwise (edge) and triplet (face) relationships leads to severe optimization difficulty. Decoupling these features allows the model to learn specialized representations for different topologies.

\begin{revblock}

\begin{figure}[t]
  \centering
  \includegraphics[width=0.9\linewidth]{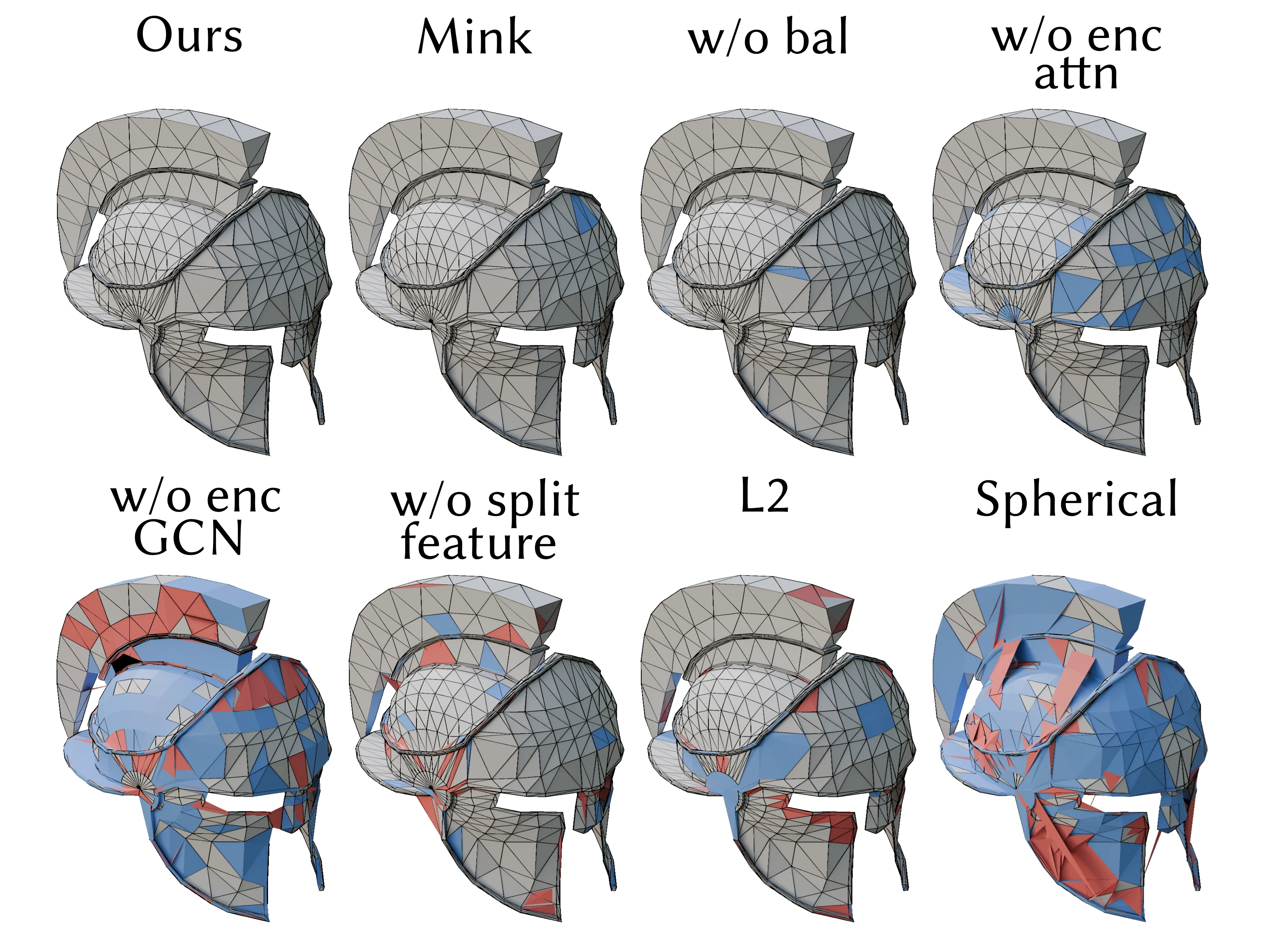}
  \vspace{-6pt}
  \caption{\rev{\textbf{Qualitative ablation comparison with zoom-in details.} Each column shows a different ablation variant. Zoom-in regions highlight topological defects (holes, floating faces, broken edges) not fully captured by F1 scores alone.}}
  \Description{ablation visual}
  \label{fig:ablation_visual}
  \vspace{-6pt}
\end{figure}

%\paragraph{Qualitative Ablation.} 
Fig.~\ref{fig:ablation_visual} provides qualitative comparisons for key ablation variants. The Euclidean ($L_2$) variant produces severe topological artifacts with many broken edges and missing faces. The Spherical variant shows even worse degradation, with large portions of connectivity lost. Removing graph convolutions (w/o enc GCN) leads to chaotic topology. In contrast, our full model achieves the best reconstruction

\end{revblock}

\begin{revblock}
\subsection{Applications}

\begin{figure}[t]
  \centering
  \includegraphics[width=\linewidth]{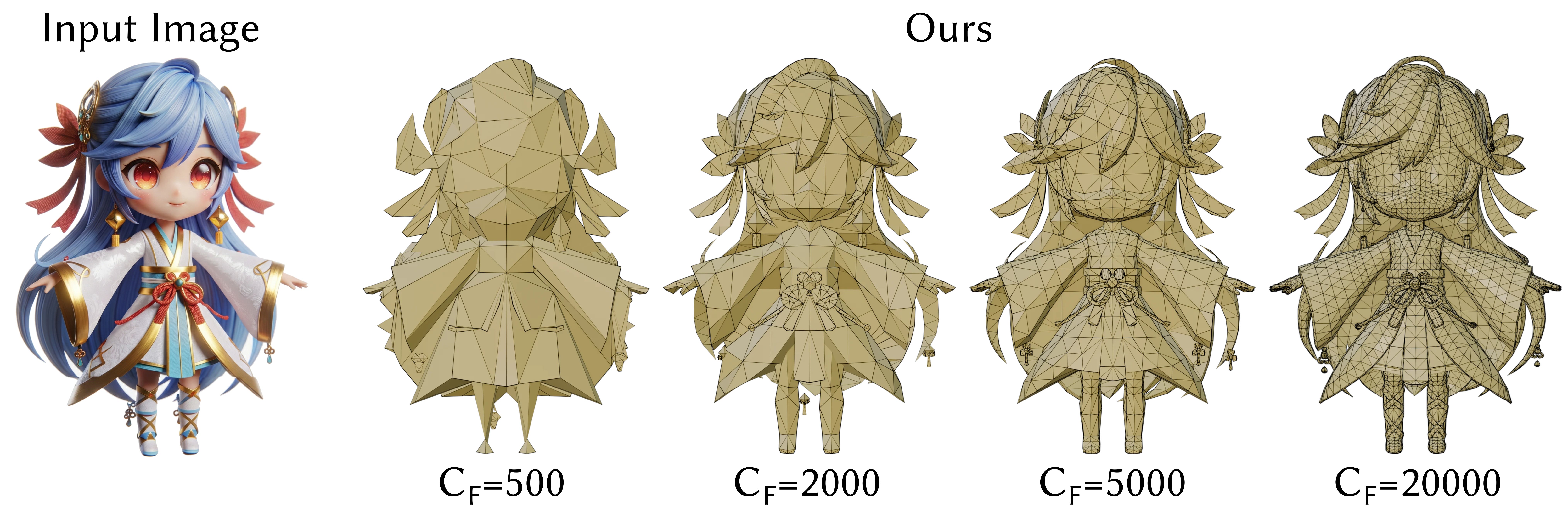}
  \vspace{-6pt}
  \caption{\rev{\textbf{Level-of-Detail control.} Given the same input, \sysname generates meshes at different face budgets (500--20{,}000), preserving global shape while varying local detail.}}
  \Description{lod}
  \label{fig:lod}
  \vspace{-6pt}
\end{figure}

\paragraph{Level-of-Detail Control.} By incorporating face-count conditioning in the first stage, \sysname can generate meshes at varying detail levels from the same input. Fig.~\ref{fig:lod} demonstrates generation at target budgets ranging from 500 to 20{,}000 faces, preserving overall shape while progressively adding geometric detail.

\begin{figure}[t]
  \centering
  \includegraphics[width=\linewidth]{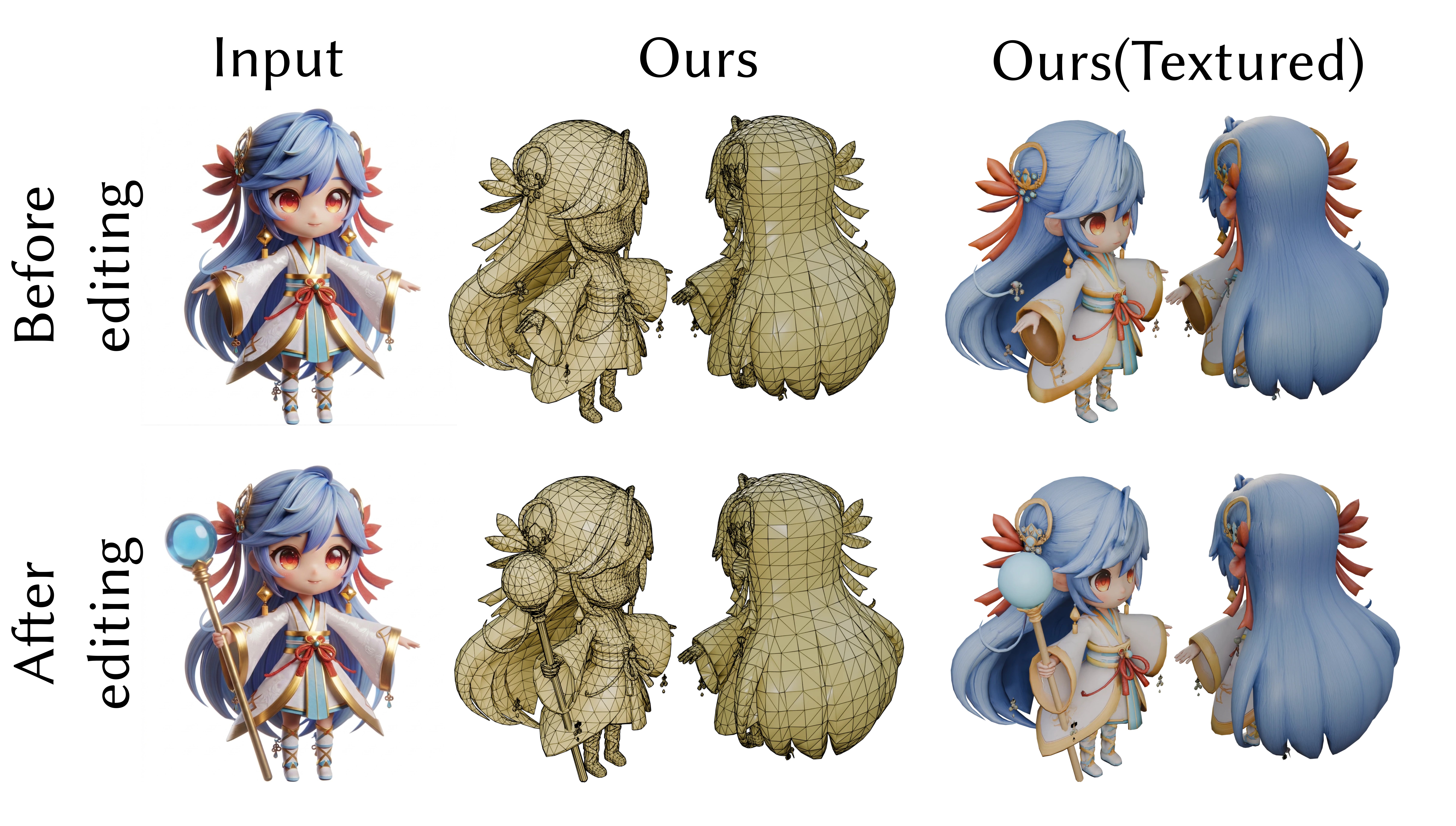}
  \vspace{-6pt}
  \caption{\rev{\textbf{Downstream applications.} \sysname meshes support texture synthesis and interactive mesh editing.}}
  \Description{editing}
  \label{fig:editing}
  \vspace{-6pt}
\end{figure}

\paragraph{Mesh Editing and Texturing.} Fig.~\ref{fig:editing} demonstrates that \sysname-generated meshes integrate seamlessly into standard graphics pipelines, supporting texture generation and interactive local mesh editing.

\begin{figure}[t]
  \centering
  \includegraphics[width=\linewidth]{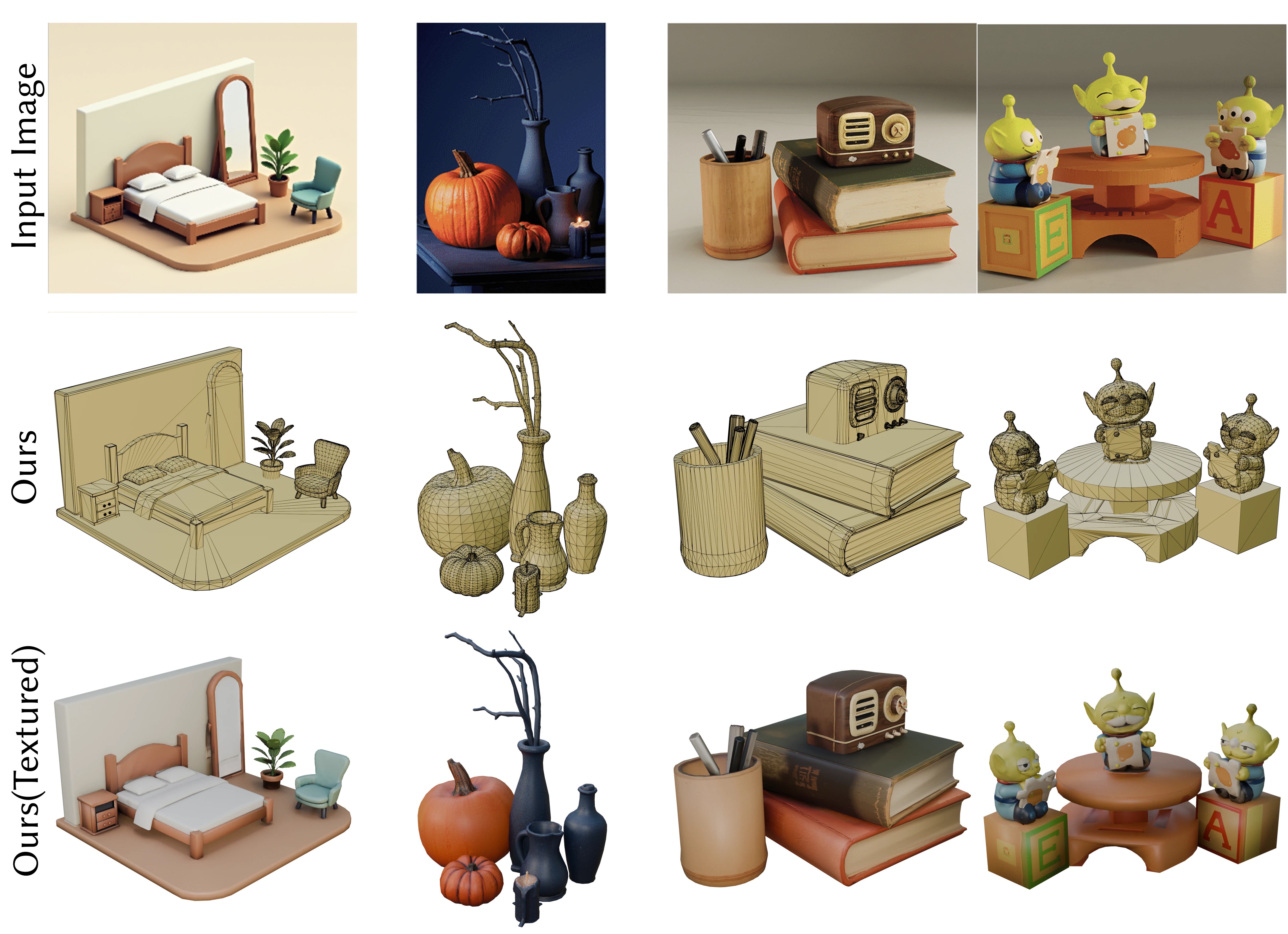}
  \vspace{-6pt}
  \caption{\rev{\textbf{Scene-level mesh generation.} By scaling to 20{,}000 faces, \sysname generates complex scenes composed of multiple objects.}}
  \Description{scene}
  \label{fig:scene}
  \vspace{-6pt}
\end{figure}

\paragraph{Scene-level Generation.} By scaling \sysname to 20{,}000 faces, we find that it can generate complex scene-level meshes composed of multiple objects (Fig.~\ref{fig:scene}). Our topology diffusion uses approximately $|\mathcal{F}|/2$ tokens compared to $\sim 9|\mathcal{F}|$ tokens for vanilla autoregressive approaches, making large-scale generation feasible with inference completing in approximately 60 seconds on a single 24GB GPU.

\end{revblock}

\section{Conclusion}

In this paper, we presented \sysname, a novel sort-free approach for holistic mesh generation that \rev{fundamentally departs from} the prevailing sequential serialization paradigm. Unlike autoregressive approaches that are plagued by permutation sensitivity and error accumulation, our method treats mesh generation as an order-agnostic, holistic process. We achieved this through several core innovations: a hierarchical octree diffusion model that synthesizes vertices in a coarse-to-fine manner, and the novel Spacetime Interval for topology representation. Based on the representation, we managed to encode discrete topology into continuous, diffusable vertex embeddings that can be easily modeled by standard diffusion models.

Extensive experiments demonstrate that \sysname significantly outperforms state-of-the-art autoregressive and two-stage baselines. \rev{A blind user study with 1{,}221 pairwise comparisons from 3D practitioners confirms strong perceptual preference for our results. Furthermore, our method demonstrates robust performance under varying input noise and point cloud density, and scales to meshes with up to 20{,}000 faces.} By decoupling geometry and topology into efficient, parallelizable diffusion processes, we not only improve generation quality but also offer a new perspective on structured mesh generation. Our work suggests that complex, non-Euclidean structures can be effectively modeled through continuous diffusion, paving the way for more robust and scalable 3D generative models.

\begin{figure}[t]
  \centering
  \includegraphics[width=\linewidth]{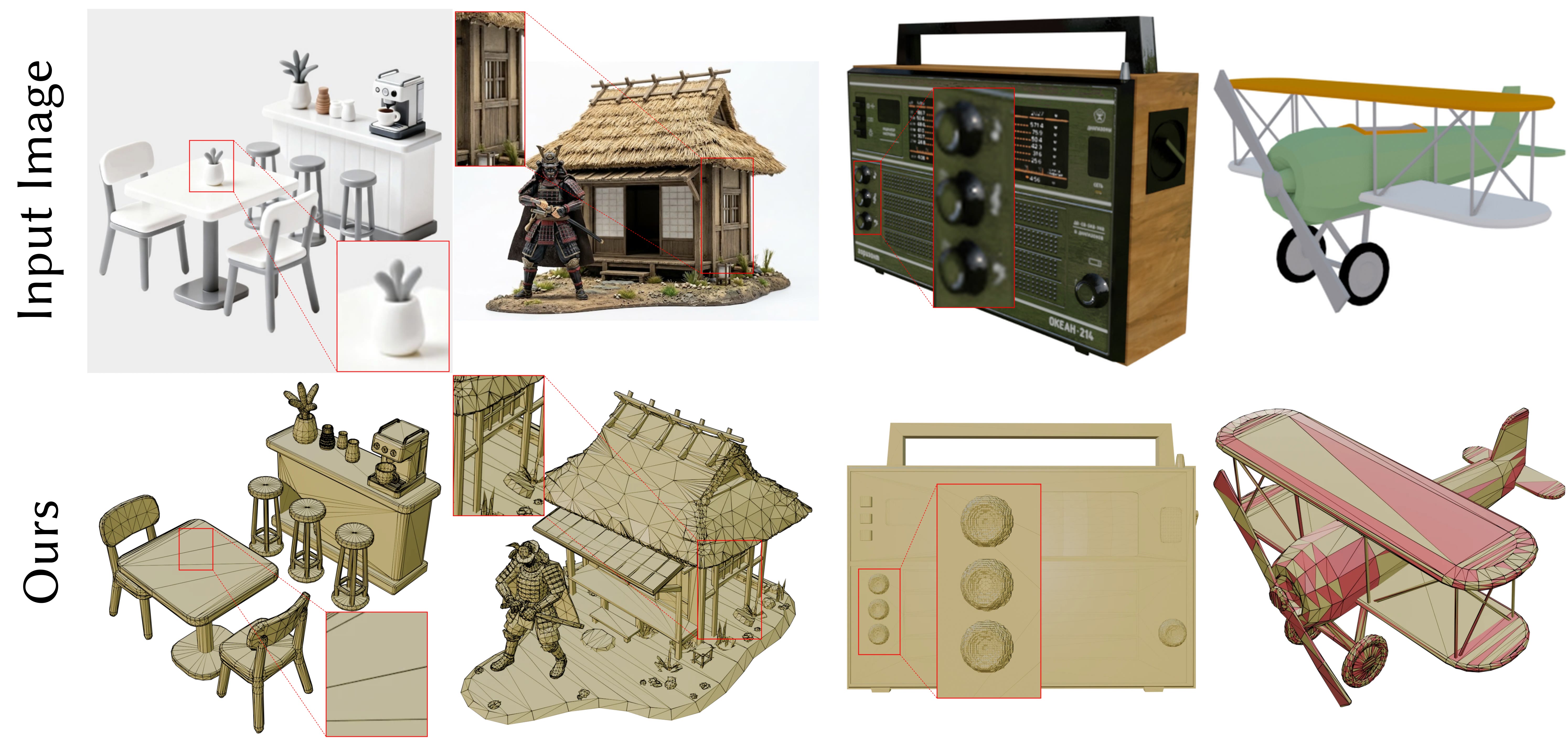}
  \vspace{-6pt}
\caption{\rev{\textbf{Failure cases.} From left to right: 
(a) \textbf{Ambiguity of geometry and texture}: missing small-scale components (e.g., a cup on the table); 
(b) \textbf{Missing faces}: large missing surfaces on building walls; 
(c) \textbf{Resolution-related artifacts}: distortions on fine structures at depth $D{=}9$ due to discrete coordinate snapping; 
(d) \textbf{Normal inconsistencies}: conflicting face orientations before the application of our orientation correction module.}}
\Description{Four subfigures showing: a missing cup due to resolution, large gaps on a wall, grid-snapping artifacts at depth 9, and inconsistent face normals.}
  \label{fig:failure}
  \vspace{-6pt}
\end{figure}

\paragraph{Limitations.} Although our approach can generate meshes in a holistic manner, the first stage can still cost around 1 minute to generate vertices in 512 resolution as it involves multiple diffusion inferences. We believe it can be accelerated by integrating few-step training or distillation techniques. In terms of the topology generation, our approach ignores the face normals so an extra orientation correction stage is required to make them consistent. We believe that a learning-based or rendering-based solution can be good enough. \rev{Also, despite of the robustness with in-the-wild inputs, it can still fails in some cases (Fig.~\ref{fig:failure}), such as ambiguous about geometry and texture, missing faces and grid artifacts with limited resolution.}

\section{Acknowledgments}
This work was supported in part by the International (Hong Kong, Macao, and Taiwan) Collaborative R\&D Project, Beijing Major Science and Technology Project under Contract No. Z251100007125016. We would like to sincerely thank Yumeng Li and Peng-Shuai Wang for their valuable suggestions and insightful discussions.

\bibliographystyle{ACM-Reference-Format}
\bibliography{bib}

\end{document}

%% file: macros.tex
\usepackage{xspace}
\def\sysname{{Nexus}\xspace}
\definecolor{revpink}{RGB}{219, 48, 122}
\definecolor{revblue}{RGB}{0, 0, 0}
\newcommand{\rev}[1]{\textcolor{revblue}{#1}}
\newenvironment{revblock}{\begingroup\color{revblue}}{\endgroup}

\usepackage{multirow}
\usepackage{enumitem}

\usepackage{pifont}% http://ctan.org/pkg/pifont
\definecolor{softblue}{RGB}{106, 156, 194}
\definecolor{softred}{RGB}{209, 125, 125}

%% file: bib.bib
@inproceedings{nash2020polygen,
  title={Polygen: An autoregressive generative model of 3d meshes},
  author={Nash, Charlie and Ganin, Yaroslav and Eslami, SM Ali and Battaglia, Peter},
  booktitle={International conference on machine learning},
  pages={7220--7229},
  year={2020},
  organization={PMLR}
}

@article{wang2017cnn,
	title={O-cnn: Octree-based convolutional neural networks for 3d shape analysis},
	author={Wang, Peng-Shuai and Liu, Yang and Guo, Yu-Xiao and Sun, Chun-Yu and Tong, Xin},
	journal={ACM Transactions on Graphics (TOG)},
	volume={36},
	number={4},
	pages={72},
	year={2017},
	publisher={ACM}
}

@inproceedings{wang2018adaptive,
  title={Adaptive O-CNN: a patch-based deep representation of 3D shapes},
  author={Wang, Peng-Shuai and Sun, Chun-Yu and Liu, Yang and Tong, Xin},
  booktitle={SIGGRAPH Asia 2018 Technical Papers},
  pages={217},
  year={2018},
  organization={ACM}
}

@inproceedings{siddiqui2024meshgpt,
  title={Meshgpt: Generating triangle meshes with decoder-only transformers},
  author={Siddiqui, Yawar and Alliegro, Antonio and Artemov, Alexey and Tommasi, Tatiana and Sirigatti, Daniele and Rosov, Vladislav and Dai, Angela and Nie{\ss}ner, Matthias},
  booktitle={Proceedings of the IEEE/CVF conference on computer vision and pattern recognition},
  pages={19615--19625},
  year={2024}
}

@article{he2025meshcraft,
  title={Meshcraft: Exploring efficient and controllable mesh generation with flow-based dits},
  author={He, Xianglong and Chen, Junyi and Huang, Di and Liu, Zexiang and Huang, Xiaoshui and Ouyang, Wanli and Yuan, Chun and Li, Yangguang},
  journal={arXiv preprint arXiv:2503.23022},
  year={2025}
}

@article{chen2024meshxl,
  title={Meshxl: Neural coordinate field for generative 3d foundation models},
  author={Chen, Sijin and Chen, Xin and Pang, Anqi and Zeng, Xianfang and Cheng, Wei and Fu, Yijun and Yin, Fukun and Wang, Billzb and Yu, Jingyi and Yu, Gang and others},
  journal={Advances in Neural Information Processing Systems},
  volume={37},
  pages={97141--97166},
  year={2024}
}

@article{hao2024meshtron,
  title={Meshtron: High-fidelity, artist-like 3d mesh generation at scale},
  author={Hao, Zekun and Romero, David W and Lin, Tsung-Yi and Liu, Ming-Yu},
  journal={arXiv preprint arXiv:2412.09548},
  year={2024}
}

@inproceedings{lionar2025treemeshgpt,
  title={Treemeshgpt: Artistic mesh generation with autoregressive tree sequencing},
  author={Lionar, Stefan and Liang, Jiabin and Lee, Gim Hee},
  booktitle={Proceedings of the Computer Vision and Pattern Recognition Conference},
  pages={26608--26617},
  year={2025}
}

@article{chen2024meshanything,
  title={Meshanything: Artist-created mesh generation with autoregressive transformers},
  author={Chen, Yiwen and He, Tong and Huang, Di and Ye, Weicai and Chen, Sijin and Tang, Jiaxiang and Chen, Xin and Cai, Zhongang and Yang, Lei and Yu, Gang and others},
  journal={arXiv preprint arXiv:2406.10163},
  year={2024}
}

@inproceedings{chen2025meshanything,
  title={Meshanything v2: Artist-created mesh generation with adjacent mesh tokenization},
  author={Chen, Yiwen and Wang, Yikai and Luo, Yihao and Wang, Zhengyi and Chen, Zilong and Zhu, Jun and Zhang, Chi and Lin, Guosheng},
  booktitle={Proceedings of the IEEE/CVF International Conference on Computer Vision},
  pages={13922--13931},
  year={2025}
}

@inproceedings{weng2025scaling,
  title={Scaling mesh generation via compressive tokenization},
  author={Weng, Haohan and Zhao, Zibo and Lei, Biwen and Yang, Xianghui and Liu, Jian and Lai, Zeqiang and Chen, Zhuo and Liu, Yuhong and Jiang, Jie and Guo, Chunchao and others},
  booktitle={Proceedings of the Computer Vision and Pattern Recognition Conference},
  pages={11093--11103},
  year={2025}
}

@inproceedings{shen2024spacemesh,
  title={Spacemesh: A continuous representation for learning manifold surface meshes},
  author={Shen, Tianchang and Li, Zhaoshuo and Law, Marc and Atzmon, Matan and Fidler, Sanja and Lucas, James and Gao, Jun and Sharp, Nicholas},
  booktitle={SIGGRAPH Asia 2024 Conference Papers},
  pages={1--11},
  year={2024}
}

@article{tang2024edgerunner,
  title={Edgerunner: Auto-regressive auto-encoder for artistic mesh generation},
  author={Tang, Jiaxiang and Li, Zhaoshuo and Hao, Zekun and Liu, Xian and Zeng, Gang and Liu, Ming-Yu and Zhang, Qinsheng},
  journal={arXiv preprint arXiv:2409.18114},
  year={2024}
}

@inproceedings{wei2025octgpt,
  title={Octgpt: Octree-based multiscale autoregressive models for 3d shape generation},
  author={Wei, Si-Tong and Wang, Rui-Huan and Zhou, Chuan-Zhi and Chen, Baoquan and Wang, Peng-Shuai},
  booktitle={Proceedings of the Special Interest Group on Computer Graphics and Interactive Techniques Conference Conference Papers},
  pages={1--11},
  year={2025}
}

@article{clay,
  title={Clay: A controllable large-scale generative model for creating high-quality 3d assets},
  author={Zhang, Longwen and Wang, Ziyu and Zhang, Qixuan and Qiu, Qiwei and Pang, Anqi and Jiang, Haoran and Yang, Wei and Xu, Lan and Yu, Jingyi},
  journal={ACM Transactions on Graphics (TOG)},
  volume={43},
  number={4},
  pages={1--20},
  year={2024},
  publisher={ACM New York, NY, USA}
}

@article{triposg,
  title={Triposg: High-fidelity 3d shape synthesis using large-scale rectified flow models},
  author={Li, Yangguang and Zou, Zi-Xin and Liu, Zexiang and Wang, Dehu and Liang, Yuan and Yu, Zhipeng and Liu, Xingchao and Guo, Yuan-Chen and Liang, Ding and Ouyang, Wanli and others},
  journal={arXiv preprint arXiv:2502.06608},
  year={2025}
}

@article{lattice,
  title={LATTICE: Democratize High-Fidelity 3D Generation at Scale},
  author={Lai, Zeqiang and Zhao, Yunfei and Zhao, Zibo and Liu, Haolin and Lin, Qingxiang and Huang, Jingwei and Guo, Chunchao and Yue, Xiangyu},
  journal={arXiv preprint arXiv:2512.03052},
  year={2025}
}

@inproceedings{trellis,
  title={Structured 3d latents for scalable and versatile 3d generation},
  author={Xiang, Jianfeng and Lv, Zelong and Xu, Sicheng and Deng, Yu and Wang, Ruicheng and Zhang, Bowen and Chen, Dong and Tong, Xin and Yang, Jiaolong},
  booktitle={Proceedings of the Computer Vision and Pattern Recognition Conference},
  pages={21469--21480},
  year={2025}
}

@misc{kim2025fastmesh,
      title={FastMesh: Efficient Artistic Mesh Generation via Component Decoupling}, 
      author={Jeonghwan Kim and Yushi Lan and Armando Fortes and Yongwei Chen and Xingang Pan},
      year={2025},
      eprint={2508.19188},
      archivePrefix={arXiv},
      url={https://arxiv.org/abs/2508.19188}, 
}

@article {Wang2023OctFormer,
     title      = {OctFormer: Octree-based Transformers for {3D} Point Clouds},
     author     = {Wang, Peng-Shuai},
     journal    = {ACM Transactions on Graphics (SIGGRAPH)},
     volume     = {42},
     number     = {4},
     year       = {2023},
 }

@misc{liu2024point,
     title={Point Mamba: A Novel Point Cloud Backbone Based on State Space Model with Octree-Based Ordering Strategy}, 
     author={Jiuming Liu and Ruiji Yu and Yian Wang and Yu Zheng and Tianchen Deng and Weicai Ye and Hesheng Wang},
     year={2024},
     eprint={2403.06467},
     archivePrefix={arXiv},
     primaryClass={cs.CV}
}

@article{ren2024octree,
  title={Octree-gs: Towards consistent real-time rendering with lod-structured 3d gaussians},
  author={Ren, Kerui and Jiang, Lihan and Lu, Tao and Yu, Mulin and Xu, Linning and Ni, Zhangkai and Dai, Bo},
  journal={arXiv preprint arXiv:2403.17898},
  year={2024}
}

@misc{guo2025hyper3defficient3drepresentation,
      title={Hyper3D: Efficient 3D Representation via Hybrid Triplane and Octree Feature for Enhanced 3D Shape Variational Auto-Encoders}, 
      author={Jingyu Guo and Sensen Gao and Jia-Wang Bian and Wanhu Sun and Heliang Zheng and Rongfei Jia and Mingming Gong},
      year={2025},
      eprint={2503.10403},
      archivePrefix={arXiv},
      primaryClass={cs.CV},
      url={https://arxiv.org/abs/2503.10403}, 
}

@misc{deng2025efficientautoregressiveshapegeneration,
      title={Efficient Autoregressive Shape Generation via Octree-Based Adaptive Tokenization}, 
      author={Kangle Deng and Hsueh-Ti Derek Liu and Yiheng Zhu and Xiaoxia Sun and Chong Shang and Kiran Bhat and Deva Ramanan and Jun-Yan Zhu and Maneesh Agrawala and Tinghui Zhou},
      year={2025},
      eprint={2504.02817},
      archivePrefix={arXiv},
      primaryClass={cs.CV},
      url={https://arxiv.org/abs/2504.02817}, 
}

@article{Xiong_2025_SGP,
  journal = {Computer Graphics Forum},
  title = {{OctFusion: Octree-based Diffusion Models for 3D Shape Generation}},
  author = {Xiong, Bojun and Wei, Si-Tong and Zheng, Xin-Yang and Cao, Yan-Pei and Lian, Zhouhui and Wang, Peng-Shuai},
  year = {2025},
  publisher = {The Eurographics Association and John Wiley & Sons Ltd.},
  ISSN = {1467-8659},
  DOI = {10.1111/cgf.70198}
}

@misc{alliegro2023polydiffgenerating3dpolygonal,
      title={PolyDiff: Generating 3D Polygonal Meshes with Diffusion Models}, 
      author={Antonio Alliegro and Yawar Siddiqui and Tatiana Tommasi and Matthias Nießner},
      year={2023},
      eprint={2312.11417},
      archivePrefix={arXiv},
      primaryClass={cs.CV},
      url={https://arxiv.org/abs/2312.11417}, 
}

@misc{zhang20233dshape2vecset3dshaperepresentation,
      title={3DShape2VecSet: A 3D Shape Representation for Neural Fields and Generative Diffusion Models}, 
      author={Biao Zhang and Jiapeng Tang and Matthias Niessner and Peter Wonka},
      year={2023},
      eprint={2301.11445},
      archivePrefix={arXiv},
      primaryClass={cs.CV},
      url={https://arxiv.org/abs/2301.11445}, 
}

@misc{vae,
      title={Auto-Encoding Variational Bayes}, 
      author={Diederik P Kingma and Max Welling},
      year={2022},
      eprint={1312.6114},
      archivePrefix={arXiv},
      primaryClass={stat.ML},
      url={https://arxiv.org/abs/1312.6114}, 
}

@misc{dinov3,
      title={DINOv3}, 
      author={Oriane Siméoni and Huy V. Vo and Maximilian Seitzer and Federico Baldassarre and Maxime Oquab and Cijo Jose and Vasil Khalidov and Marc Szafraniec and Seungeun Yi and Michaël Ramamonjisoa and Francisco Massa and Daniel Haziza and Luca Wehrstedt and Jianyuan Wang and Timothée Darcet and Théo Moutakanni and Leonel Sentana and Claire Roberts and Andrea Vedaldi and Jamie Tolan and John Brandt and Camille Couprie and Julien Mairal and Hervé Jégou and Patrick Labatut and Piotr Bojanowski},
      year={2025},
      eprint={2508.10104},
      archivePrefix={arXiv},
      primaryClass={cs.CV},
      url={https://arxiv.org/abs/2508.10104}, 
}

@misc{deitke2022objaverseuniverseannotated3d,
      title={Objaverse: A Universe of Annotated 3D Objects}, 
      author={Matt Deitke and Dustin Schwenk and Jordi Salvador and Luca Weihs and Oscar Michel and Eli VanderBilt and Ludwig Schmidt and Kiana Ehsani and Aniruddha Kembhavi and Ali Farhadi},
      year={2022},
      eprint={2212.08051},
      archivePrefix={arXiv},
      primaryClass={cs.CV},
      url={https://arxiv.org/abs/2212.08051}, 
}

@misc{deitke2023objaversexluniverse10m3d,
      title={Objaverse-XL: A Universe of 10M+ 3D Objects}, 
      author={Matt Deitke and Ruoshi Liu and Matthew Wallingford and Huong Ngo and Oscar Michel and Aditya Kusupati and Alan Fan and Christian Laforte and Vikram Voleti and Samir Yitzhak Gadre and Eli VanderBilt and Aniruddha Kembhavi and Carl Vondrick and Georgia Gkioxari and Kiana Ehsani and Ludwig Schmidt and Ali Farhadi},
      year={2023},
      eprint={2307.05663},
      archivePrefix={arXiv},
      primaryClass={cs.CV},
      url={https://arxiv.org/abs/2307.05663}, 
}

@misc{toys4k,
      title={Using Shape to Categorize: Low-Shot Learning with an Explicit Shape Bias}, 
      author={Stefan Stojanov and Anh Thai and James M. Rehg},
      year={2021},
      eprint={2101.07296},
      archivePrefix={arXiv},
      primaryClass={cs.CV},
      url={https://arxiv.org/abs/2101.07296}, 
}

@misc{sglang,
      title={SGLang: Efficient Execution of Structured Language Model Programs}, 
      author={Lianmin Zheng and Liangsheng Yin and Zhiqiang Xie and Chuyue Sun and Jeff Huang and Cody Hao Yu and Shiyi Cao and Christos Kozyrakis and Ion Stoica and Joseph E. Gonzalez and Clark Barrett and Ying Sheng},
      year={2024},
      eprint={2312.07104},
      archivePrefix={arXiv},
      primaryClass={cs.AI},
      url={https://arxiv.org/abs/2312.07104}, 
}

@inproceedings{vllm,
  title={Efficient Memory Management for Large Language Model Serving with PagedAttention},
  author={Woosuk Kwon and Zhuohan Li and Siyuan Zhuang and Ying Sheng and Lianmin Zheng and Cody Hao Yu and Joseph E. Gonzalez and Hao Zhang and Ion Stoica},
  booktitle={Proceedings of the ACM SIGOPS 29th Symposium on Operating Systems Principles},
  year={2023}
}

@article{octree,
  title={Geometric modeling using octree encoding},
  author={Meagher, Donald},
  journal={Computer graphics and image processing},
  volume={19},
  number={2},
  pages={129--147},
  year={1982},
  publisher={Elsevier}
}

@misc{rope,
      title={RoFormer: Enhanced Transformer with Rotary Position Embedding}, 
      author={Jianlin Su and Yu Lu and Shengfeng Pan and Ahmed Murtadha and Bo Wen and Yunfeng Liu},
      year={2023},
      eprint={2104.09864},
      archivePrefix={arXiv},
      primaryClass={cs.CL},
      url={https://arxiv.org/abs/2104.09864}, 
}

@misc{dit,
      title={Scalable Diffusion Models with Transformers}, 
      author={William Peebles and Saining Xie},
      year={2023},
      eprint={2212.09748},
      archivePrefix={arXiv},
      primaryClass={cs.CV},
      url={https://arxiv.org/abs/2212.09748}, 
}

@misc{dpm-solver,
      title={DPM-Solver: A Fast ODE Solver for Diffusion Probabilistic Model Sampling in Around 10 Steps}, 
      author={Cheng Lu and Yuhao Zhou and Fan Bao and Jianfei Chen and Chongxuan Li and Jun Zhu},
      year={2022},
      eprint={2206.00927},
      archivePrefix={arXiv},
      primaryClass={cs.LG},
      url={https://arxiv.org/abs/2206.00927}, 
}

@inproceedings{flowmatching,
  author       = {Yaron Lipman and
                  Ricky T. Q. Chen and
                  Heli Ben{-}Hamu and
                  Maximilian Nickel and
                  Matthew Le},
  title        = {Flow Matching for Generative Modeling},
  booktitle    = {{ICLR}},
  publisher    = {OpenReview.net},
  year         = {2023}
}

@inproceedings{rectifiedflow,
  author       = {Xingchao Liu and
                  Chengyue Gong and
                  Qiang Liu},
  title        = {Flow Straight and Fast: Learning to Generate and Transfer Data with
                  Rectified Flow},
  booktitle    = {{ICLR}},
  publisher    = {OpenReview.net},
  year         = {2023}
}

@article{DBLP:journals/corr/HamiltonYL17,
  author       = {William L. Hamilton and
                  Rex Ying and
                  Jure Leskovec},
  title        = {Inductive Representation Learning on Large Graphs},
  journal      = {CoRR},
  volume       = {abs/1706.02216},
  year         = {2017},
  url          = {http://arxiv.org/abs/1706.02216},
  eprinttype   = {arXiv},
  eprint       = {1706.02216},
  bibsource    = {dblp computer science bibliography, https://dblp.org}
}

@misc{lai2025latticedemocratizehighfidelity3d,
      title={LATTICE: Democratize High-Fidelity 3D Generation at Scale}, 
      author={Zeqiang Lai and Yunfei Zhao and Zibo Zhao and Haolin Liu and Qingxiang Lin and Jingwei Huang and Chunchao Guo and Xiangyu Yue},
      year={2025},
      eprint={2512.03052},
      archivePrefix={arXiv},
      primaryClass={cs.GR},
      url={https://arxiv.org/abs/2512.03052}, 
}

@misc{ulip,
      title={ULIP: Learning a Unified Representation of Language, Images, and Point Clouds for 3D Understanding}, 
      author={Le Xue and Mingfei Gao and Chen Xing and Roberto Martín-Martín and Jiajun Wu and Caiming Xiong and Ran Xu and Juan Carlos Niebles and Silvio Savarese},
      year={2023},
      eprint={2212.05171},
      archivePrefix={arXiv},
      primaryClass={cs.CV},
      url={https://arxiv.org/abs/2212.05171}, 
}

@misc{uni3d,
      title={Uni3D: Exploring Unified 3D Representation at Scale}, 
      author={Junsheng Zhou and Jinsheng Wang and Baorui Ma and Yu-Shen Liu and Tiejun Huang and Xinlong Wang},
      year={2023},
      eprint={2310.06773},
      archivePrefix={arXiv},
      primaryClass={cs.CV},
      url={https://arxiv.org/abs/2310.06773}, 
}
